\documentclass{article}

\usepackage{microtype}
\usepackage{graphicx}
\usepackage{subfigure}
\usepackage{booktabs} 

\usepackage{hyperref}

\usepackage[accepted]{icml2023}

\usepackage{amsmath}
\usepackage{amssymb}
\usepackage{mathtools}
\usepackage{amsthm}

\usepackage{booktabs}
\usepackage{multirow}
\usepackage{threeparttable}
\usepackage{multicol}
\usepackage{bm}
\usepackage{amsfonts} 
\usepackage{comment}

\usepackage[capitalize,noabbrev]{cleveref}

\theoremstyle{plain}

\theoremstyle{definition}

\theoremstyle{remark}

\usepackage[textsize=tiny]{todonotes}

\icmltitlerunning{Rethink DARTS Search Space and Renovate a New Benchmark}

\begin{document}

\twocolumn[
\icmltitle{Rethink DARTS Search Space and Renovate a New Benchmark}



\icmlsetsymbol{equal}{*}

\begin{icmlauthorlist}
\icmlauthor{Jiuling Zhang}{yyy,comp}
\icmlauthor{Zhiming Ding}{comp,yyy}
\end{icmlauthorlist}

\icmlaffiliation{yyy}{University of Chinese Academy of Sciences, Beijing, China}
\icmlaffiliation{comp}{Institute of Software, Chinese Academy of Sciences, Beijing, China}

\icmlcorrespondingauthor{Zhiming Ding}{zhiming@iscas.ac.cn, zhangjiuling19@mails.ucas.edu.cn}

\icmlkeywords{AutoDL, Neural Architecture Search, Differentiable Architecture search}

\vskip 0.3in
]



\printAffiliationsAndNotice{}  

\begin{abstract}
DARTS search space (DSS) has become a canonical benchmark for NAS whereas some emerging works pointed out the issue of narrow accuracy range and claimed it would hurt the method ranking. We observe some recent studies already suffer from this issue that overshadows the meaning of scores. In this work, we first propose and orchestrate a suite of improvements to frame a larger and harder DSS, termed LHD, while retaining high efficiency in search. We step forward to renovate a LHD-based new benchmark, taking care of both discernibility and accessibility. Specifically, we re-implement twelve baselines and evaluate them across twelve conditions by combining two underexpolored influential factors: transductive robustness and discretization policy, to reasonably construct a benchmark upon multi-condition evaluation. Considering that the tabular benchmarks are always insufficient to adequately evaluate the methods of neural architecture search (NAS), our work can serve as a crucial basis for the future progress of NAS. \href{https://github.com/chaoji90/LHD}{https://github.com/chaoji90/LHD}
\end{abstract}

\section{Introduction}
DARTS relaxes categorical selection through a convex combination of the architecture parameters and operation outputs. In the search phase, architecture parameters $\alpha$ and operation weights $\omega$ are alternately optimized on validation set and training set respectively through a bilevel optimization objective. Henceforth, we collectively refer to the line of works explicitly parameterize architecture search by relaxing the categorical operation selection to a differentiable operation distribution as DARTS and specify the method pioneered by \citet{liu2018darts} as vanilla DARTS. We also use the name of search space to refer to the benchmark on that space when the context is unambiguous. Research community has established a benchmark surrounding DSS which has been extensively used to evaluate NAS methods \cite{mehta2021bench}.
Given a current benchmark, two desiderata are discernibility and accessibility.
\citet{li2020random} studied the indeterministic training of the methods and demonstrated that empirically, validation accuracy (\textit{val\_acc}) fluctuates over multiple trials, sometimes exceeding 0.1\%, for the same finalnet (search result) under the same seed. Accordingly, we refer to the case where the  accuracy margin in rank is less than 0.1\% as  \textbf{n}arrow \textbf{r}ange \textbf{r}anking (NRR). Table~\ref{table1} illustrates current scores on DSS where the average \textbf{a}ccuracy \textbf{m}argin between \textbf{a}djacent items of the method \textbf{r}anking (AMAR) is only 0.012\% on CIFAR-10 which is 8.3$\times$ smaller than 0.1\%. Some studies \cite{yang2019evaluation,garg2020revisiting,yu2019evaluating,wang2020neural} have pointed out that the narrow  accuracy range of DSS causes baselines indiscernible and impairs the validity of the benchmark (more related works in Appx.A).

\begin{table}[t]
\caption{Scores of the benchmark on DSS.}
\label{table1}
\begin{center}
\begin{small}
\begin{sc}
\resizebox{0.49\textwidth}{!}{
\begin{tabular}{lccc}
\toprule
\multirow{2}{*}{\textbf{Methods}} & \multicolumn{2}{c}{\textbf{CIFAR-10}} & \multirow{2}{*}{\textbf{ImageNet-1K}} \\
                         & Error (\%)        & \textit{\#param} (M)    &                              \\
\midrule 
MiLeNAS \cite{he2020milenas}                 & 2.51±0.11    & 3.87          & 24.7                         \\
PC-DARTS \cite{xu2019pc}                & 2.57±0.07    & 3.6           & 25.1                         \\
GAEA-ERM \cite{li2021geometry}           & 2.50±0.06    & 3.7           & 24.3                         \\
DrNAS \cite{chen2021drnas}                   & 2.54±0.03    & 4.0           & 24.2                         \\
GibbsNAS \cite{xue2021rethinking}                & 2.53±0.02    & 4.1           & 24.6                         \\
SP-DARTS  \cite{zhang2021robustifying}               & 2.50±0.07    & 3.5           & 24.4                         \\
DARTS- \cite{chu2020darts}& 2.59±0.08&3.5&24.8\\
$\beta$-DARTS \cite{ye2022b}& 2.53±0.08&3.83&24.2\\
\bottomrule
\end{tabular}
}
\end{sc}
\end{small}
\end{center}
\vskip -0.1in
\end{table}

\citet{yang2019evaluation} systematically studied the evaluation phase on DSS and incrementally quantified the contribution of different modules (Auxiliary Towers, Drop Path, Cutout, Channels, AutoAugment, Epochs) to the final scores. They emphasized that introducing new tricks in \textit{evaluation} has a much greater impact on performance than employing different NAS methods. By contrast to their work focused on the manifest influential factors, we observe more subtleties, i.e. several minute deviations of the evaluation protocol are introduced by succeeding studies, including different drop-path rate, learning rate decay target, batch size, seed, minor revise of operations (batch normalization after poolings). These minutiae are partially inherited by subsequent researches \cite{xue2021rethinking,li2021geometry,zhang2021robustifying,chen2021drnas} but are intractable unless carefully investigating every released code. We combine the above minutiae and re-evaluate the finalnet (search result) reported by vanilla DARTS and obtain $+$0.14\% (2.76$\to$2.62) improvement which is much larger than 0.012\% and gives us reason to believe that the cumulative effect of these modifications must be non-trivial in light of the NRR of Table~\ref{table1}.

To sum up, \textbf{margin} in rank is critical for the confidence of a benchmark. For this, we find some previous studies use \textit{t-test} to measure the confidence of the score comparisons \cite{hooker2019compressed,yu2019evaluating,pourchot2020share}.
So in this paper, we utilize both AMAR and the average \textit{t-test} \cite{welch1947generalization} margin between adjacent items of ranking (TMAR) as the measurements of discernibility. For a list of methods $L$, we can i) Sort $L$ in terms of accuracies (rank); ii) Get pair-wise margins (accuracy gap for AMAR, \textit{t-test} for TMAR) of adjacent items of the sorted $L$; iii) Average all margins to get AMAR or TMAR.
AMAR measures the absolute margin of accuracies and TMAR takes both accuracy and variance into account when examining the \textit{t-test} value.
We can also see from the 4\textit{th} column in Table~\ref{table1} that the narrow range is likely to be an intrinsic feature of the search space and will not be rectified by simply evaluating on more challenging data. This observation is also verified on other datasets by \citet{yang2019evaluation}.

In general, a good space of NAS is expected to exclude human bias and be flexible enough to cover a wide variety of candidates \cite{he2021automl}. Most of the previous search spaces were proposed as the \textit{byproducts of methods}. To challenge the art scores, these studies always have an incentive to introduce as many artifacts as possible to make their space easier to traverse so that the methods are less error-prone. However, \textit{this design motive is diametrically opposite to the discernible objective of benchmark thereby leads to the problem of NRR we are observing now}. On the contrary, AutoML-Zero \citep{real2020automl} specifically pointed out that the  accuracy of a large enough search space should be sparse which is very the critical character of a discernible benchmark. Even further, too many artifacts also cause the search space easy to be overfitted. \citet{he2020milenas} observed that the parameter scale (\textit{\#param}) is closely related to  \textit{val\_acc} and outperforms art zero-shot estimators on DSS \cite{ning2021evaluating}. FLOPs and \textit{\#param} remain highly correlated and exhibits consistent correlation with \textit{val\_acc} on both DSS and our LHD, so in this paper we focus on inspecting \textit{\#param}. We believe that the approach to get better score on benchmark by just looking for larger capacity operations is definitely not our expectation for NAS. The space of a benchmark should be both \textbf{large} and \textbf{difficult}, so that the methods are not prone to attain higher scores by opportunistically overfitting the space. In this work, we propose some improvements to overhaul DSS and formulate a \textbf{l}arger and \textbf{h}arder new \textbf{D}SS, namely LHD. 

Based on LHD, we step ahead to newly construct a multi-condition evaluation benchmark in which we focus on combining the evaluation of both transductive robustness and discretization policies. The `transductive' here refers to search and expect to find the optimal architecture in-situ on the search dataset. The benefits of the multi-condition evaluation is three-fold: i) Further enhance discriminability, even if some methods perform close under a single condition, we can compare them by taking all conditions into account; ii) Make benchmark more challenging, claiming superior across multiple conditions is much harder than on a single condition; iii) Uncover many more methodological characteristics and preferences that are unobservable within current counterpart that solely provides a few scores.
Our contributions can be summarized as follows:

\textbf{1.} We propose i) Node aggregation enhancement: input-softmax; ii) New searchable blocks: searchable polynary operations, searchable cell outputs of sum and concatenation; iii) Primitive refinement: unified convolution primitive. We orchestrate all to construct a new search space that is demonstrably larger and harder than DSS. Through this work, we succeed in weakening the correlation between \textit{\#param} and \textit{val\_acc} from 0.52 (KD $\tau$) on DSS ~\cite{ning2021evaluating,yang2019evaluation} to 0.29/0.26/0.20 on three valid spaces of LHD.

\textbf{2.} We renovate a new benchmark on LHD involves assessing the transductive robustness of \textbf{twelve} baselines over four discretization policies across three datasets. No single method outperforms others under all conditions and the overall ranks are rather unstable across conditions both of which demonstrate that our benchmark is more challenging for methods to show generalizability and claim superior.

\section{New Search Space}
\textbf{Preliminary}: DSS formulates a cell-based search space with $N$ nodes  $X = \{ {x^n}|n \in [1..N]\} $ where $[1..N] := [1,N]$ and $E$ compound edges $G = \{ g_{{i,j}}^e|e \in [1..E],i \in [1,j - 1],j \in ({n_i},N]\} $ where $n_i$ refers to the number of cell inputs. Compound edge ${g_{i,j}}$ connects node $i$ to $j$ and associates three attributes: candidate operation set ${O_{i,j}} = \{ o_{{i,j}}^m|m \in [1..M]\} $, corresponding operation parameter set ${A_{i,j}} = \left\{ {\alpha _{i,j}^m|m \in [1..M]} \right\}$, probability distribution of the parameters ${\bm{a}_{i,j}} = {\rm{softmax}} ({A_{i,j}})$. Every intermediate node is connected to all its predecessor through an edge ${g_{i,j}}({x_i}) =  \left\langle {\bm{a}_{i,j}},{O_{i,j}}({x_i}) \right\rangle $ where $i < j$. Typically, a unified set of operation candidates $O = \{ {o^m}|m \in [1..M]\} $ is defined for all edges. Each edge subsumes all operation candidates to express the transformations between nodes. Every node aggregates the outputs of all incoming edges into the new feature maps. The network that encodes all architecture candidates is called supernet. We use primitive to describe an indivisible substructure of operations and denote block as a substructure containing nodes, edges, paths which can be searchable (s/e) or unsearchable (u/e). We use rounded and solid rectangle, dashed box to represent cell, block and primitive respectively.

\textbf{Design principle}: We first propose a suite of improvements and new searchable blocks. For clarity, we separately delineate their motivations, problem solved, and solutions. We then orchestrate them all to frame the new  LHD. An ablation of these improvements are provided in Section 3. Finally, we give a number of features of the new search space. For the overall design principle, \textit{we enlarge search space and trim artifacts while keeping its empirical memory cost roughly the same}. Artifacts refer to the unsearchable structures in macro design and the over-designed primitives of operation candidates, both of which introduce human bias, limit the flexibility of space, make the space easy to traverse and less error prone. We realize beforehand that some similar ideas were proposed and examined separately~\cite{wu2021neural,jiang2019improved} but not collectively. All these researches are far from framing a generally applicable new space to replace DSS.  We formulate LHD in Appx.B.

\begin{figure}[t]
\centering
\includegraphics[width=0.89\columnwidth]{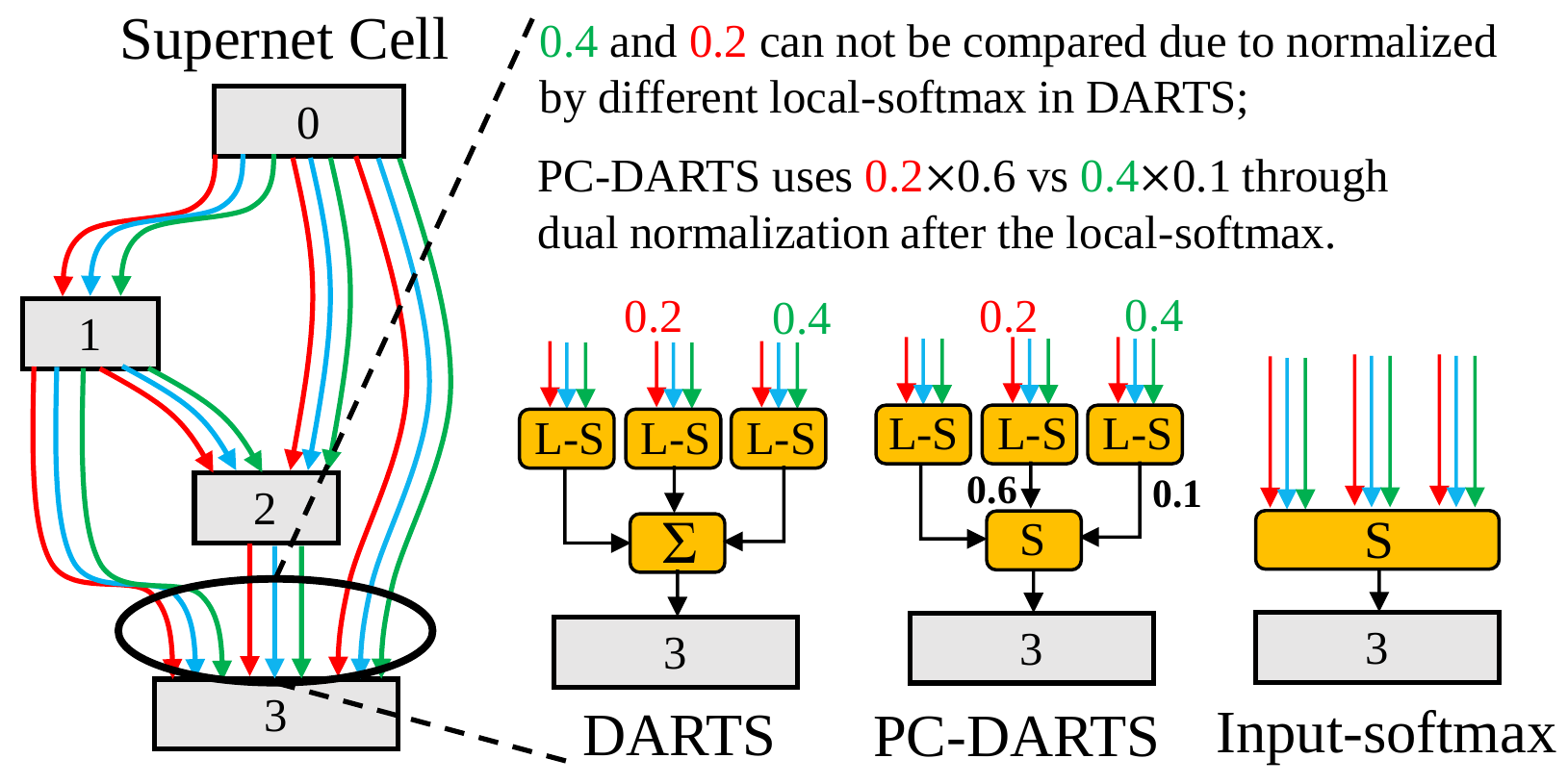} 
\caption{Zoom in on the input end of the node $3$ to illustrate the differences. L-S denotes the local-softmax in DARTS.}
\label{fig2}
\vskip 0.1in
\includegraphics[width=0.8\columnwidth]{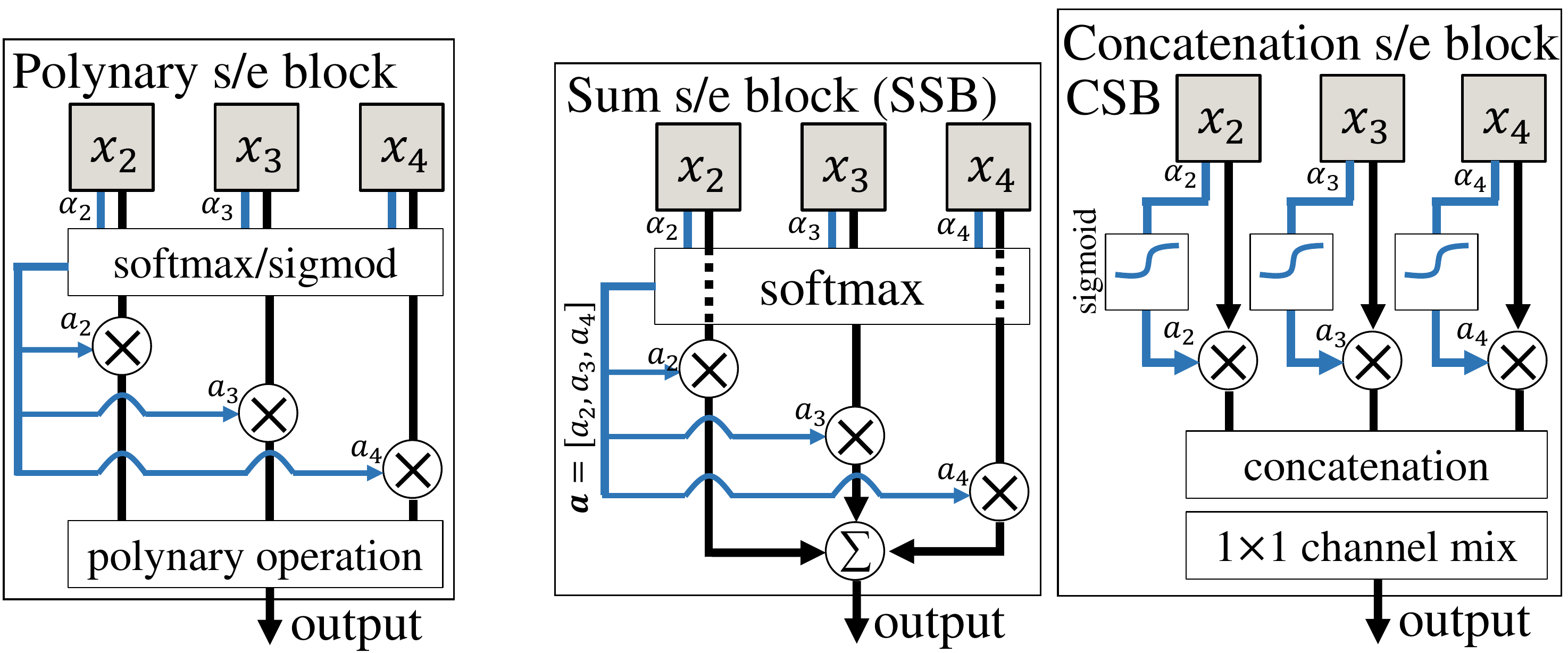} 
\caption{Polynary s/e block on the left and its two instances on the right. One parameter attach to each path to express the significance that can be optimized through SGD. Parameters are first squashed by softmax or sigmoid and then weight the feature maps passed through the paths. LHD uses s/e blocks on the right as cell outputs.}
\label{fig3}
\vskip -0.1in
\end{figure}

\subsection{New modules to frame LHD}
\textbf{Input-softmax}: \textit{Motivation: DSS suffers from the limit of sub-graphs due to the absence of a mechanism to compare the significance of operations across edges}. As shown in Figure~\ref{fig2}, the softmax is applied on each compound edge without considering their connection pattern in graph, namely local-softmax by I-DARTS \cite{jiang2019improved}.
Node aggregation in DSS is simply sums up all the feature maps of incoming edges as ${x_j} = \sum\nolimits_{i < j} {{g_{i,j}}({x_i})} $. PC-DARTS partially solves this by employing path normalization ${p_{i,j}}{g_{i,j}}$ to double normalize the significances from different edges depicted in Figure~\ref{fig2}. \textit{Solution: We address this limitation by placing softmax directly before aggregation on the node input end} instead of edges to simultaneously normalize elements across all incoming edges shown on the rightmost of Figure~\ref{fig2}. This way, the significance of any operation $o_{i,j}^m$ toward node $j$ can be fairly compared through the value of $a_{i,j}^m$ for all the combinations of $m \in [1,M]$ and $i \in [1,j - 1]$. 

\begin{figure}[t]
\centering
\includegraphics[width=1\columnwidth]{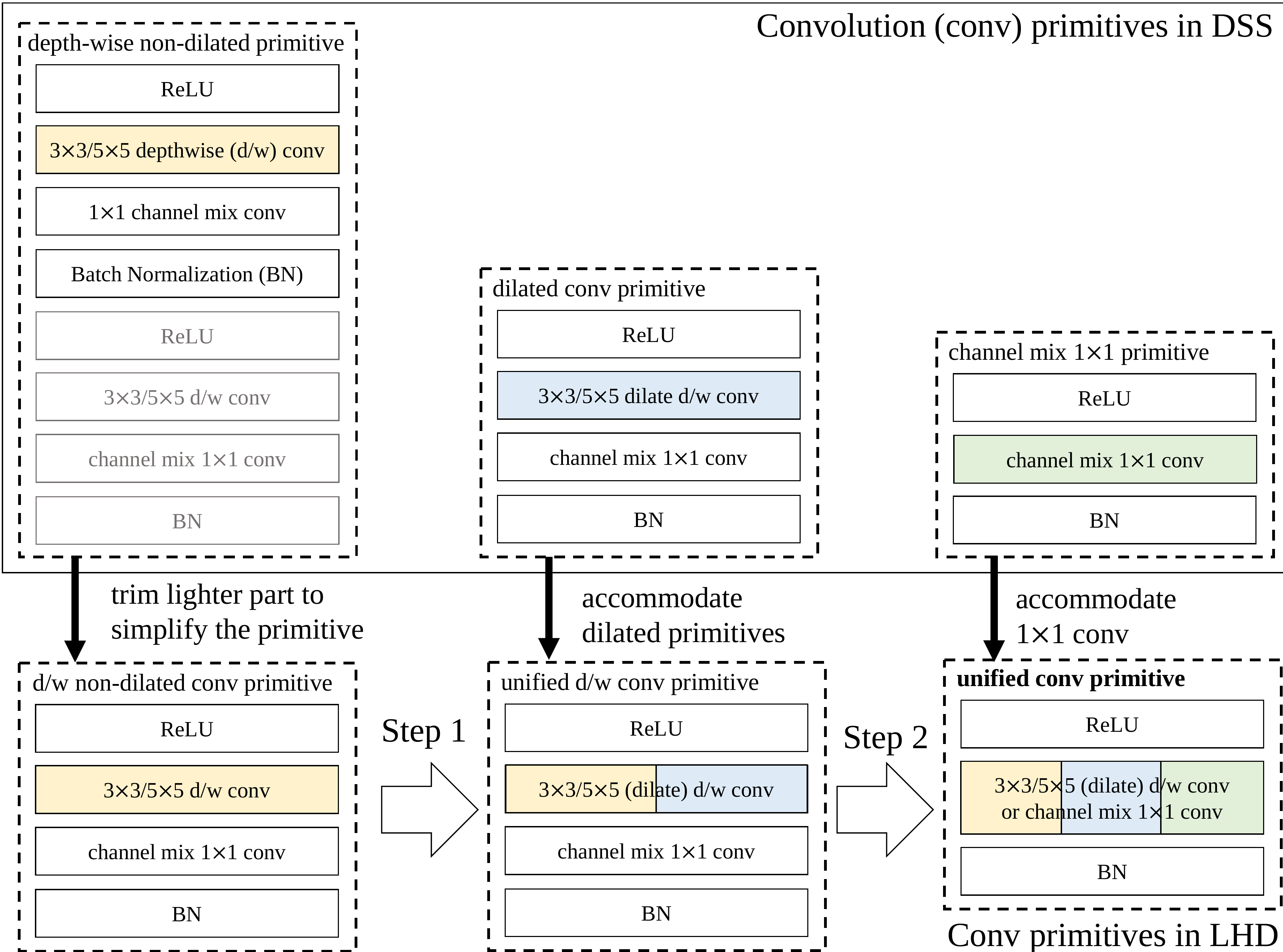} 
\vskip -0.05in
\caption{Refinement and unification aim to simplify and unify the structure of convolution primitives to trim redundant artifacts, the operations are still conducted separately not merged~\cite{wang2021mergenas}.}
\label{fig4}
\vskip 0.1in
\includegraphics[width=1.03\columnwidth]{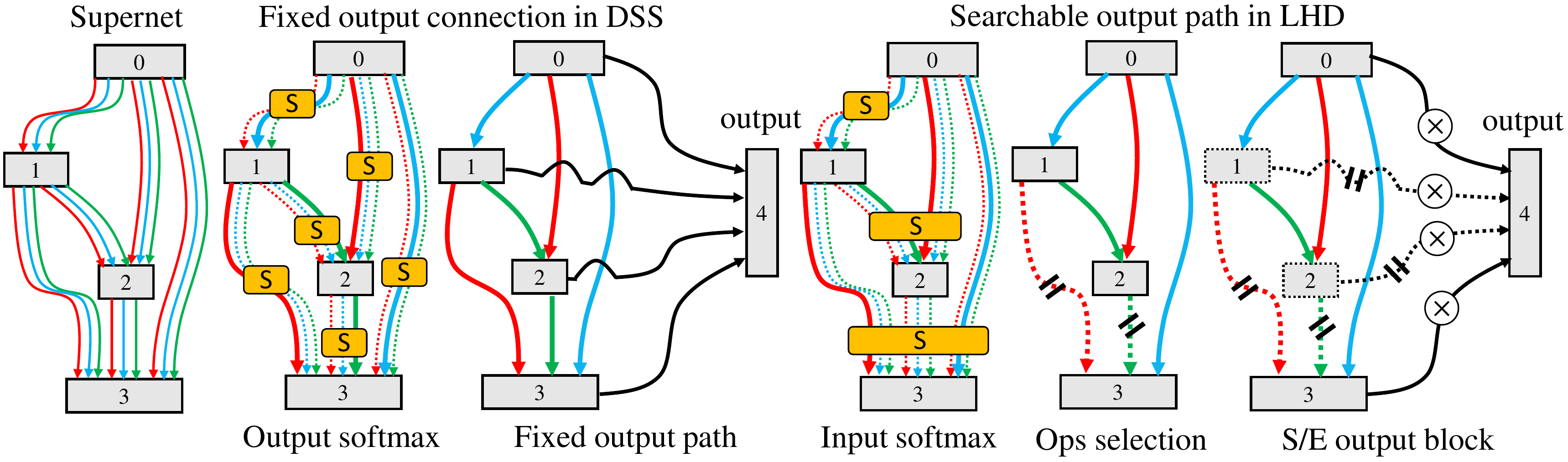}
\vskip -0.05in
\caption{Fixed output path in DSS on the left compared to s/e output block and removable intermediate nodes in LHD on the right. By orchestrating input-softmax and polynary s/e  block, the intermediate node on the rightmost can be removed from the finalnet (search result) in the following two cases: i) Node is neither selected by the cell output nor selected by any subsequent nodes like node $2$; ii) Node is neither selected by the cell output nor any its succeeding node is selected by the output like node $1$.}
\label{fig5}
\vskip -0.1in
\end{figure}

\textbf{Search Polynary Operation}: 
\textit{Motivation: Innovative use of the polynary operations is often a key improvement in handcrafted regime}, e.g. addition in ResNet~\cite{he2016deep} and concatenation in DenseNet~\cite{huang2017densely}.
However, parameters only attach on unary operations (single input single output) on each edge in DSS. 
\textit{Solution: Generalizing the merit of DARTS to search polynary operations is straightforward by associating parameter with each path to express its significance that can be optimized through gradient descent}. Figure~\ref{fig3} conceptually visualizes the s/e block of polynary operation on the left. The path parameters are first squashed and then weight and aggregate the feature maps over all paths.

\textbf{Unified Convolution Primitive}: 
\textit{Motivation: As shown in Figure~\ref{fig4}, the convolution primitives are designed to be rather complicated and the non-dilated primitives are deliberately deeper than the dilated counterparts in DSS}. For trimming artifacts to reduce human bias, \textit{solution: we first unify the dilated and non-dilated depth-wise (d/w) primitives specified as step 1 to form the unified d/w primitive. We step forward to incorporate the d/w primitive and 1$\times$1 channel mixer into an unified structure of the primitive}. The unified primitive ultimately accommodates all convolution candidates and ensures their similar structures in the space.

\textbf{Orchestration to Build LHD}: 
\textit{Motivation: DARTS is incapable to search none (zero) operation directly. So all intermediate nodes are densely connected to the cell output and unsearchably attend in finalnet as shown on the left of Figure~\ref{fig5}}. The valid size of space is thus severely restricted to solely depends on the number of edges $M$ because the search on DSS is limited to only the operation selection inside edges without considering their interconnection topology. \textit{Solution: We relax the connection path between intermediate nodes and cell output and let methods search the inter-cell connection pattern through optimizing the path significance in the search phase}. Our goal is to make the intermediate nodes \textbf{removable} thereby decouple the finalnet from the design of supernet in Figure~\ref{fig5}.

\begin{figure}[t]
\centering
\includegraphics[width=1\columnwidth]{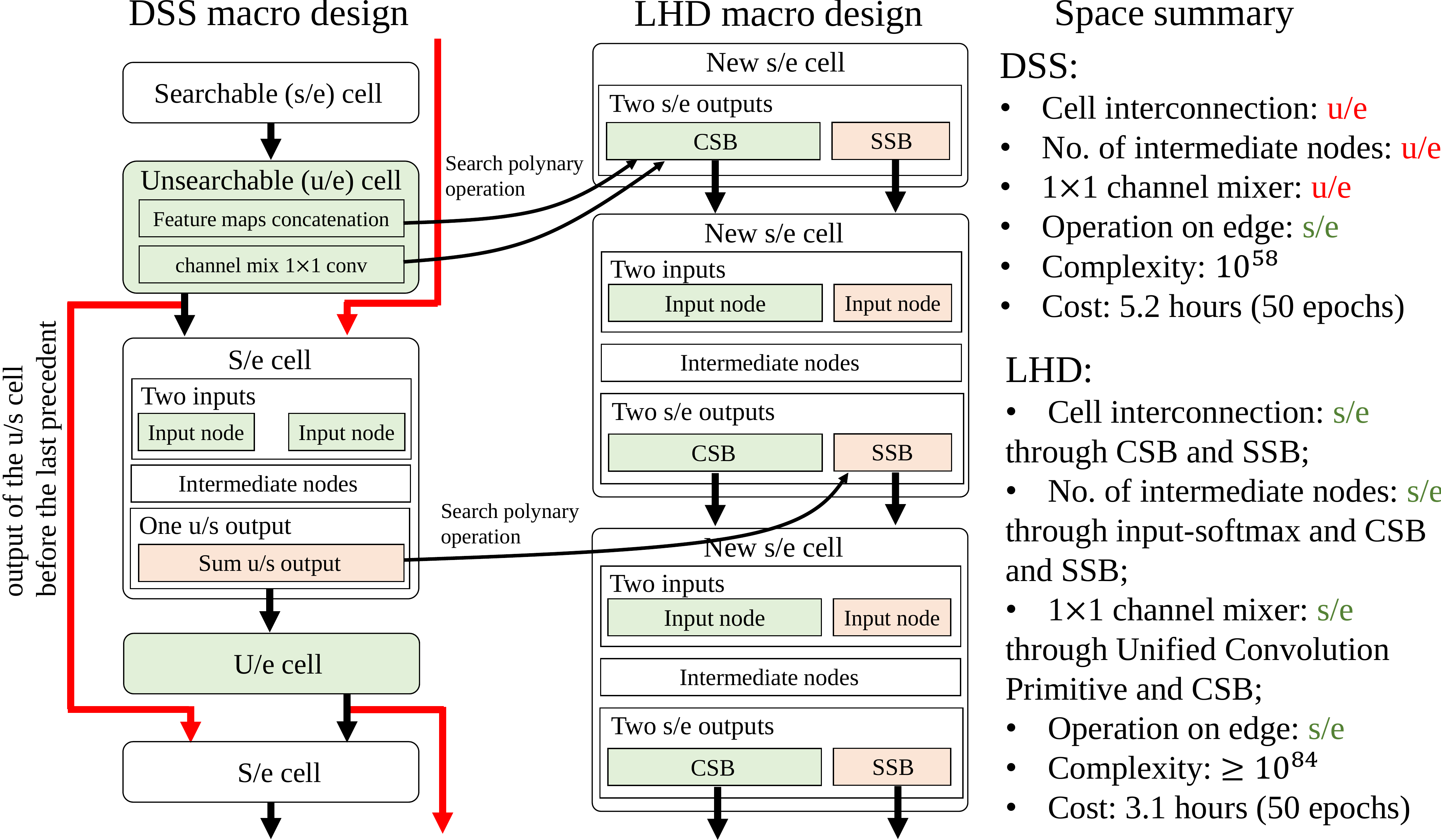} 
\vskip -0.05in
\caption{Macro design of DSS involves two major artifacts: i) u/e cell for concatenation and channel mix; ii) fixed skip connections between cells. We trim artifact `i' by instantiating a polynary s/e block as a concatenation s/e block (CSB) and use CSB as an output of s/e cell to replace the u/e cell in DSS as shown by the black arrow. We then trim `ii' but retain dual inputs of the s/e cell and instantiate another sum s/e block (SSB) as the second output to match the number of inputs of s/e cell as shown on the right.}.
\label{fig6}
\vskip 0.05in
\includegraphics[width=0.75\columnwidth]{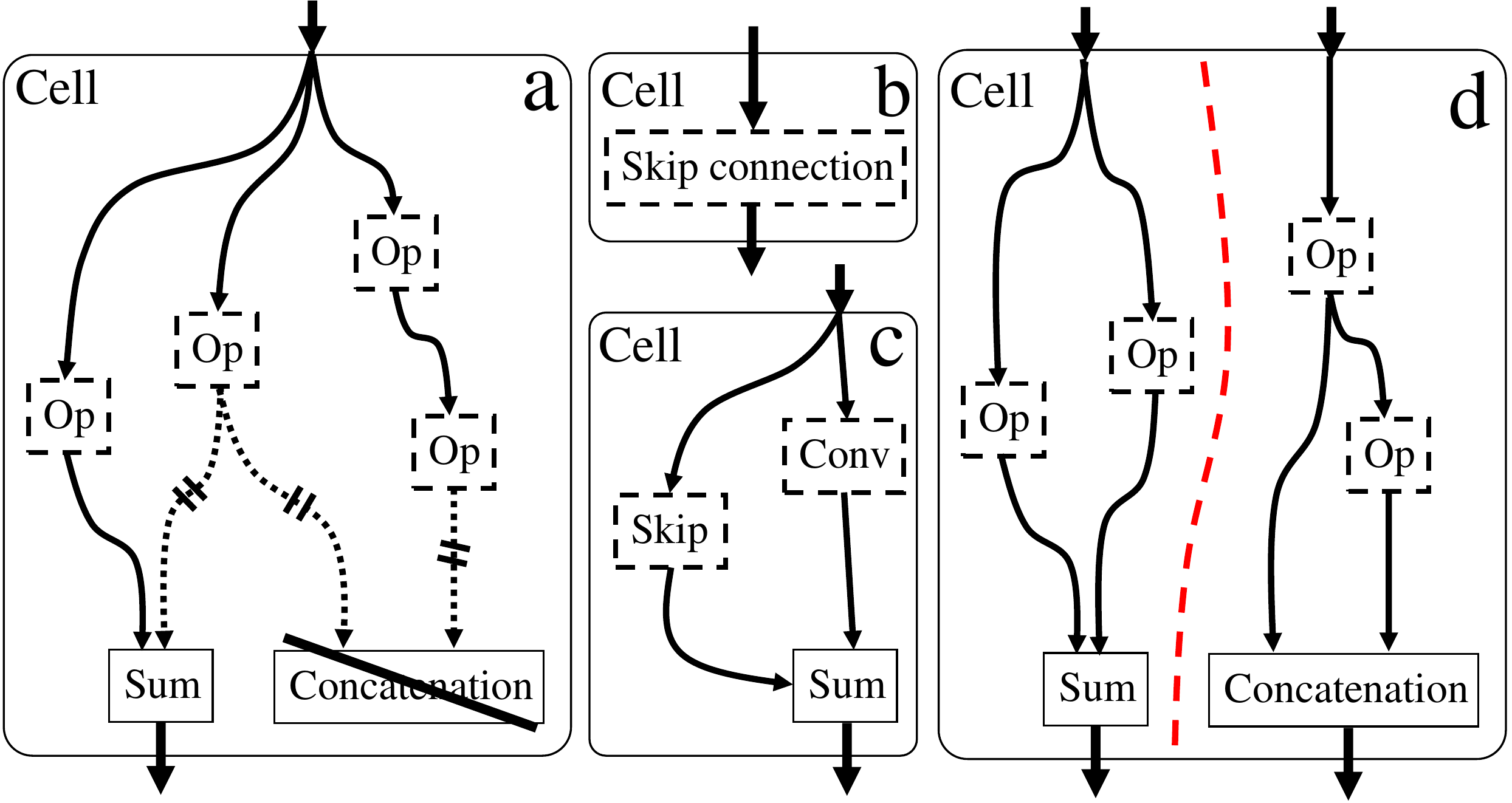} 
\caption{Three corner cases (a,b,d) accommodated by LHD but absent in DSS shows the versatility and inclusivity of LHD.}
\label{fig7}
\vskip -0.1in
\end{figure}

\textit{Motivation: Artifacts of the u/e cell and fixed skip connection in the macro design of DSS} as shown on left of Figure~\ref{fig6}. As a design choice, we regard the path of SSB as an selection with exclusivity in contrast to the non-exclusive path selection of CSB. So we normalize the output path of CSB and SSB by gating and softmax respectively as detailed on the right of Figure~\ref{fig3}. \textit{Solution: Detailed in the caption of Figure~\ref{fig6}}. By pruning the fixed inter-cell skip connections highlighted in red in Figure~\ref{fig6}, LHD expects NAS method to learn the appropriate gradient path by themselves in the search phase such that the methods cannot obtain better score by simply choosing larger capacity architecture.

\subsection{Characteristics of LHD}
\textbf{Case study}: SSB normalizes the output path by softmax allude to that at least the strongest path will be reserved. Meanwhile, the sigmoid gating in CSB have a potential to close all path that leads to finalnet reduces to a single-input single-output structure as shown in Figure~\ref{fig7}a. The cell output is a feature maps aggregation by summation that coincides with the building block of ResNet in Figure~\ref{fig7}c. In another case shown by Figure~\ref{fig7}b, SSB selects only one output path from the intermediate node that reads feature maps only from the cell input through a skip connection. The cell input and output are straight through without transformation that of course won’t bring any reasonable performance but will be a meaningful failure case. NAS methods can also find cell with two parallel branches as shown by Figure~\ref{fig7}d and the representations are thus learned independently.

\begin{table*}[t]
\centering
\caption{Outline specific characteristics of baselines in the benchmark. Penultimate column lists the benchmarks used for evaluation in original papers. NB201 is NAS-BENCH-201 \cite{dong2019bench}, NB1S1 is NAS-BENCH-1Shot1 \cite{zela2019bench}, S1$\sim$S4 are proposed by \cite{arber2020understanding}. (T) denotes the tabular benchmark. PC denotes partial channels. SP denotes sparse ${\bm{a}}$ 
 distribution and OS denotes operation shortcut. GAEA-B refers to GAEA-Bilevel and GAEA-E refers to GAEA-ERM \cite{li2021geometry}. $\beta$-DARTS added an additional term in loss to regularize architecture parameters.}
\label{table3}
\vskip 0.1in
\resizebox{\textwidth}{!}{
\begin{tabular}{lccccccccccc}
\toprule
Baseline & \multicolumn{2}{c}{Optimization} & Relaxation & Gradient & SP & PC & OS & Evaluations adopted & Codebase\\
\midrule 
DARTS \cite{liu2018darts} & bilevel & joint & softmax & normal & no & no& no & DSS & quark0/darts\\
MiLeNAS \cite{he2020milenas}& \textbf{mixlevel} & joint & softmax & normal & no & no& no & DSS & chaoyanghe/MiLeNAS\\
DrNAS \cite{chen2021drnas}& bilevel & joint & \textbf{dirichlet} & normal & no & no& no & DSS, NB201 (T) & xiangning-chen/DrNAS\\
GAEA-B \cite{li2021geometry}& bilevel & joint & softmax & \textbf{exponentiated} & no & no& no & DSS, NB201 (T) & liamcli/gaea\_release\\
GAEA-E \cite{li2021geometry}& \textbf{silevel} & joint & softmax & \textbf{exponentiated} & no & no& no & DSS, NB201 (T), NB1S1 (T) & liamcli/gaea\_release\\
GDAS \cite{dong2019searching}& bilevel & \textbf{sampling} & \textbf{gumble-softmax} & normal & \textbf{yes} & no& no & DSS, NB201 (T) & D-X-Y/AutoDL-Projects\\
SP-DARTS \cite{zhang2021robustifying}& bilevel & joint & \textbf{low-temp softmax} & normal & \textbf{yes} & no& no & DSS, NB201 (T), S1$\sim$S4 & chaoji90/SP-DARTS\\
PC-DARTS \cite{xu2019pc}& bilevel & joint & softmax & normal & no & \textbf{yes}& no & DSS&yuhuixu1993/PC-DARTS \\
SurgeNAS \cite{luo2022surgenas} & \textbf{silevel} & joint & softmax & normal & no & no& \textbf{yes} & NB201 (T) & -\\
DARTS- \cite{chu2020darts} & bilevel & joint & softmax & normal & no & no& \textbf{yes} & DSS, NB201 (T), S1$\sim$S4 & Meituan-AutoML/DARTS-\\
$\beta$-DARTS \cite{ye2022b} & bilevel & joint & softmax & normal & no & no& no & DSS, NB201 (T) & Sunshine-Ye/Beta-DARTS\\
\bottomrule
\end{tabular}
}
\vskip -0.1in
\end{table*}

\begin{table}[t]
\centering
\caption{Comparisons of discretization policies.}
\label{table4}
\vskip 0.1in
\resizebox{0.47\textwidth}{!}{
\begin{tabular}{ccccc}
\toprule
\multirow{2}{*}{\begin{tabular}[c]{@{}c@{}}Search \\space\end{tabular}} & \multirow{2}{*}{\begin{tabular}[c]{@{}c@{}}Discretization \\policy\end{tabular}} & \multirow{2}{*}{\begin{tabular}[c]{@{}c@{}}Operation \\selection\end{tabular}} & \multirow{2}{*}{\begin{tabular}[c]{@{}c@{}}Output path \\selection\end{tabular}} & \multirow{2}{*}{Complexity} \\
& & & & \\
\midrule 
DSS & original & top-2 & fixed, unsearchable & $10^{18}$ \\
\midrule 
\multirow{4}{*}{LHD} & Base & top-2 & \textbf{threshold control } & $10^{31}$  \\
 & 1M & as Base & as Base & as Base  \\
 & 3ops & \textbf{top-3} & as Base & $10^{41}$  \\
 & 4out & as Base & \textbf{top-4} & $10^{28}$ \\
\bottomrule
\end{tabular}
}
\vskip -0.1in
\end{table}

\textbf{Computation and memory overhead}:
We strike a balance between the space augmentation and the search acceleration.
For vanilla DARTS, primitive refinement reduces the size of superent by 70\% (1.93M$\to$0.56M) and gives rise to a memory surplus to increase intermediate nodes from four to five. Furthermore, replacing the u/s cell in DSS with the CSB allows us to increase the batch size of the search phase by 15\% (152$\to$176). On the whole, the depth of the supernet is reduced by two-thirds and the time overhead of the search phase of vanilla DARTS on CIFAR-10 is 40\% lower than that on DSS (5.2h$\to$3.1h on RTX 3090, like-for-like comparison after aligning all other conditions).

\textbf{Complexity of the continuous DAGs}:
We increase the complexity from ${10^{58}}$ of DSS to $\geq{10^{84}}$ of LHD. Analysis is detailed in Appx.C. The valid subspace is determined by the discretization policy actually used in Table~\ref{table4}.

\section{LHD Benchmark}
Appropriate benchmark grounds existing method and inspires further research. 
Our work is not to frame a new space and try every existing method to  bring us a good architecture. To some extent, we actually embarrass existing methods by removing their dependent artifacts, enlarging search space, searching upon different conditions. Nonetheless, we will show the potentiality of some search results in the last part.
We assess twelve baselines over four discretization policies across three standard benchmark datasets tot twelve organized conditions to validate the transductive robustness and exhibit the impact of the discretization policies on ranking. Baselines are curated with specific characteristics that are highlighted in Table~\ref{table3} from which we can see that DSS is almost compulsory for the adequate evaluation of NAS methods. We do our best to fully understand the codebases released on DSS and migrate them to LHD.

\textbf{Transductive robustness in search}: NAS aims to automate the general network design. Robustness of the search phase is critical since tuning hyperparameters on each space for each dataset is prohibitive and unsustainable. We can first reasonably assume that all baselines’ settings have been specially tuned on DSS on CIFAR-10 (DSS\&C10). We expect to achieve reasonable performance by directly applying these settings on LHD\&C10. After that, we transfer these settings of the search phase to LHD\&CIFAR-100 (C100) and LHD\&SVHN to evaluate the transductive robustness across datasets (model settings are detailed in Appx.D). In our benchmark, we adopt the latest and most intuitive search protocol from \cite{li2021geometry,dong2019bench,he2020milenas,xue2021rethinking} which can be summarized as follows: i) Uniformly sample $n$ seeds from 1 to 100,000; ii) Search $n$ times with the seeds independently; iii) Evaluate $n$ results separately and take the average \textit{val\_acc}. Our benchmark set $n$ to five, the maximum value of previous studies, greater than three trials in \cite{dong2019bench,ying2019bench} and four trials in \cite{he2020milenas,xue2021rethinking}.

\textbf{Discretization policies}: We propose four discretization policies in Table~\ref{table4}. \textbf{Base} closely follows the top-2 operation selection in DSS while \textbf{3ops} acts as a straightforward alternative to select top-3 operations on each input-softmax. For the s/e cell outputs, Base thresholds SSB by 0.2 (starting point of the five s/e paths) and CSB by the mean gating level of all paths. \textbf{4out} is a variant of this by simply selecting top-4 out of five paths for both SSB and CSB so that the cells are densely interconnected. We discuss the tuning-based path selection method~\cite{wang2020rethinking} in detail in Appx.F.

DARTS expects the gradient-based optimization to select the appropriate operation. This selection subsequently affects the network scale as shown in Table~\ref{table1} but leaves the question of whether the performance gaps come from different \textit{\#param} rather than architectural merit. \cite{tay2022scaling} also identifies that models operate well in one scale does not guarantee its performant in another. In practice, networks are widely scaled by increasing depth and width.
We therefore come up policy \textbf{1M} that adopts the same architecture as Base, but first scale it up by increasing the stacked cells from 20 to 25 and then align \textit{\#param}  through augmenting \textit{init\_channels} until the finalnet attains 1.5M \textit{\#param}.

\begin{table*}[t]
\centering
\caption{We report mean and standard deviation of \textit{val\_acc}  as the main scores. We also report the averaged \textit{\#param} (M) to uncover the preference of baselines in the perspective of model capacity. We report additional top-1 and top-3 scores because some methods are trapped in rare failure cases. Evaluation results on C10 are shown here, results on C100 and SVHN are provided in Appx.E.}
\label{table5}
\vskip 0.1in
\resizebox{\textwidth}{!}{
\begin{tabular}{l|ccc|ccc|ccc|ccc}
\toprule
C10            & \multicolumn{3}{c|}{Base}           & \multicolumn{3}{c|}{1M}             & \multicolumn{3}{c|}{3ops}            & \multicolumn{3}{c}{4out}            \\
Method         & \textit{val\_acc} (\%)   & \textit{\#param} & top-1/top3  & \textit{val\_acc} (\%)  & \textit{\#param} & top-1/top3  & \textit{val\_acc} (\%)   & \textit{\#param} & top-1/top3  & \textit{val\_acc} (\%)  & \textit{\#param} & top-1/top3  \\
\midrule
DARTS          & 93.58±4.11 & 0.67     & 96.40/96.02 & 93.89±4.42 & 1.54     & 96.92/96.41 & 92.24±4.01  & 0.74     & 95.53/94.49 & 91.73±4.82 & 0.90     & 96.54/94.44 \\
DrNAS          & 94.14±2.03 & 0.57     & 95.97/95.28 & 94.48±2.20 & 1.55     & 96.39/95.77 & 94.52±1.12 & 0.64     & 95.63/95.12 & 92.75±3.80 & 0.86     & 94.83/94.72 \\
GAEA-B         & \textbf{95.93±0.46} & 0.68     & 96.52/96.18 & \textbf{96.20±0.59} & 1.56     & 96.73/96.56 & \textbf{95.70±0.83}  & 0.77     & 96.40/96.19 & \textbf{96.17±0.43} & 0.96     & 96.64/96.40 \\
GAEA-E         & 94.74±0.56 & 0.91     & 95.25/95.08 & 94.88±0.48 & 1.60      & 95.47/95.22 & 94.97±0.53  & 1.13     & 95.32/95.25 & 94.89±0.44 & 1.03     & 95.25/95.14 \\
GDAS           & 94.90±0.22 & 0.55     & 95.24/95.04 & 95.62±0.28 & 1.53     & 96.28/95.79 & 95.24±0.42  & 0.61     & 95.58/95.53 & \textbf{95.92±0.31} & 0.98     & 96.28/96.12 \\
MiLeNAS        & \textbf{95.60±0.43} & 0.66     & 96.11/95.82 & \textbf{95.99±0.55} & 1.53     & 96.86/96.30 & \textbf{95.46±0.81}  & 0.78     & 96.31/95.94 & \textbf{95.67±0.37} & 0.98     & 96.21/95.91 \\
PC-DARTS       & 95.11±0.52 & 0.71     & 95.86/95.42 & 95.45±0.56 & 1.56     & 96.22/95.76 & 95.42±0.52  & 0.83     & 96.04/95.77 & 95.40±0.51 & 1.09     & 96.09/95.73 \\
Random         & 95.09±0.69 & 0.64     & 95.75/95.51 & 95.58±0.63 & 1.56     & 96.34/95.98 & 95.60±0.11  & 0.77     & 95.75/95.68 & 95.48±0.29 & 0.92     & 95.79/95.67 \\
SP-DARTS       & \textbf{95.83±0.39} & 0.64     & 96.22/96.08 & \textbf{96.12±0.52} & 1.55     & 96.84/96.43 & \textbf{95.53±0.78}  & 0.76     & 96.22/96.07 & 94.10±2.18 & 0.94     & 96.01/95.40 \\
DARTS-         & 92.84±2.77 & 0.66     & 96.48/94.56 & 93.14±2.72 & 1.54     & 96.98/94.70 & 93.57±1.91  & 0.75     & 96.28/94.76 & 91.69±2.52 & 0.84     & 95.85/93.01 \\
$\beta$--DARTS & 95.01±0.75 & 0.60     & 95.84/95.26 & 95.16±0.98 & 1.52     & 96.03/95.47 & 93.80±1.14  & 0.71     & 94.95/94.19 & 92.11±4.47 & 0.92     & 95.22/94.26 \\
SurgeNAS       & 94.38±0.99 & 0.78     & 95.07/94.95 & 94.66±1.01 & 1.55     & 95.60/95.36 & 94.77±1.24  & 0.96     & 95.40/95.33 & 94.64±0.56 & 0.98     & 95.27/94.97\\
\bottomrule
\end{tabular}
}
\vskip -0.05in
\caption{Discernibility measurements of the ranks under different conditions. Larger values of AMAR and TMAR in LHD imply greater margins between items in rank, thereafter yield more discernible ranking to alleviate NRR in DSS.}
\label{table6}
\vskip 0.09in
\resizebox{\textwidth}{!}{
\begin{tabular}{lccccccccccccc}
\toprule
Condition & C10\&DSS & C10\&Base & C10\&1M & C10\&3ops & C10\&4out & C100\&Base & C100\&1M & C100\&3ops & C100\&4out & SVHN\&Base & SVHN\&1M & SVHN\&3ops & SVHN\&4out \\
\midrule
AMAR/\_top3 (\%) & 0.012/0.005 & 0.29/0.17 & 0.28/0.10 & 0.32/0.09 & 0.41/0.25 & 1.60/0.74 & 2.18/1.03 & 1.12/1.12 & 2.45/1.01 & 0.27/0.017 & 0.27/0.027 & 0.20/0.078 & 0.31/0.114 \\
TMAR/\_top3 & 0.30/0.08 & 0.50/0.64 & 0.43/0.30 & 0.43/0.24 & 0.67/1.07 & 0.78/0.24 & 0.60/0.43 & 0.72/0.45 & 1.12/0.77 & 0.79/0.18 & 0.98/0.25 & 0.96/1.30 & 1.18/1.47 \\
\bottomrule
\end{tabular}
}
\vskip -0.05in
\end{table*}

\begin{figure*}[!htbp]
\centering
\includegraphics[width=1\textwidth]{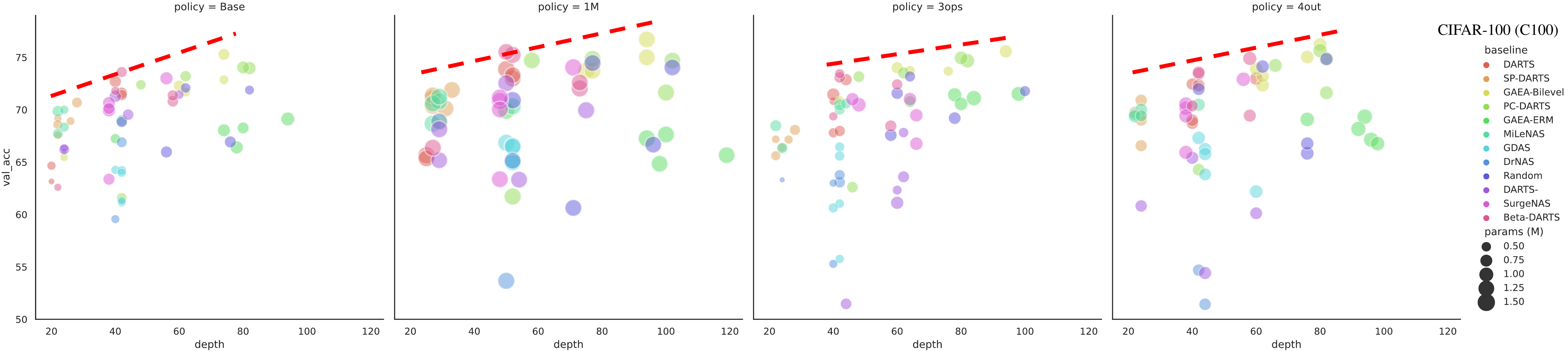}
\vskip -0.1in
\caption{Distributions of the search results on the coordinate frame of \textit{val\_acc} versus depth. We specifically put the results on C100 here because the performance ceiling is noticeably positively correlated with the finalnet (search result) depth, which is not obvious on C10 and SVHN (see Appx.E). Methods prefer deeper or shallower structures thereby struggle to achieve consistent scores across conditions.}
\label{fig8}
\vskip 0.05in
\includegraphics[width=1\textwidth]{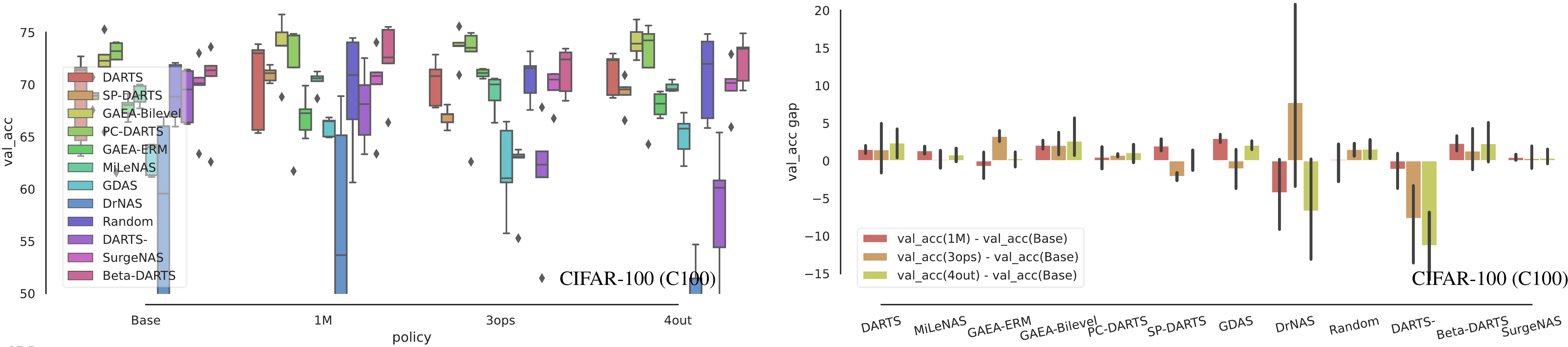} 
\vskip -0.1in
\caption{(left) Compare baseline \textit{val\_acc}s on four policies respectively along x axis. Some whiskers are truncated for the clarity in the main scope. (right) Exhibits the differences in \textit{val\_acc}s between Base and other policies to illustrate the effect of the discretization on different methods. Each bar contains the scores of five trials. Results on C10 and SVHN and are provided in Appx.E due to space limit.}
\label{fig9}
\vskip -0.1in
\end{figure*}

\textbf{Complexity of the valid search space}: Original policy of DSS allows $\prod _{{\rm{k = 1}}}^4C_{k + 1}^2 \times {7^2} = \prod _{{\rm{k = 1}}}^4\frac{{\left( {k + 1} \right)k}}{2} \times {7^2} \approx {10^9}$ possible valid cell after discretization and the total complexity of the normal and reduction searches is approximate ${10^{18}}$. Possible discretized valid cell in LHD can be obtained through ${({(\sum\nolimits_{i = 1}^{N - 1} {C_N^i} )^2}\prod\nolimits_{j = 2}^{N + 1} {C_{Mj}^k} )^2}$ with $M$ operation candidates and $N$ intermediate nodes. $k$ is the specified top-$k$ operation selection on each input-softmax. 

\textbf{Contribution ablation}: From the complexity of $10^{18}$ on DSS, contribution of each improvement can be ablated: i) For $N$=$4$ and fixed path single cell output, input-softmax enlarges valid search space by one order of magnitude to $10^{19}$; ii) Combining dual cell s/e outputs with input-softmax to yield removable intermediate node augments another six orders of magnitude from step `i' to the valid search space $10^{25}$; iii) Increasing intermediate nodes from four to five accounts for another six orders of magnitude to the final valid complexity $10^{31}$ of Base as shown in Table~\ref{table4}.

\textbf{Evaluation protocol}: To refrain the additional cost added by multi-condition evaluation, we choose a relatively lower \textit{\#param} regime ($\le$1M or 1.5M) than DSS ($\geq$3.5M). We employ seed $0$ which is the same as DSS for all evaluations on LHD and propose a $i$-value based heuristic regularization protocol (more in Appx.G) to tackle the diversity of search results.
Nevertheless, we believe that our benchmark is friendly and accessible to any practitioner since the entire pipeline can be delivered on a single GPU.

\subsection{Results of the Benchmark}
\begin{figure}[t]
\centering
\includegraphics[width=0.82\columnwidth]{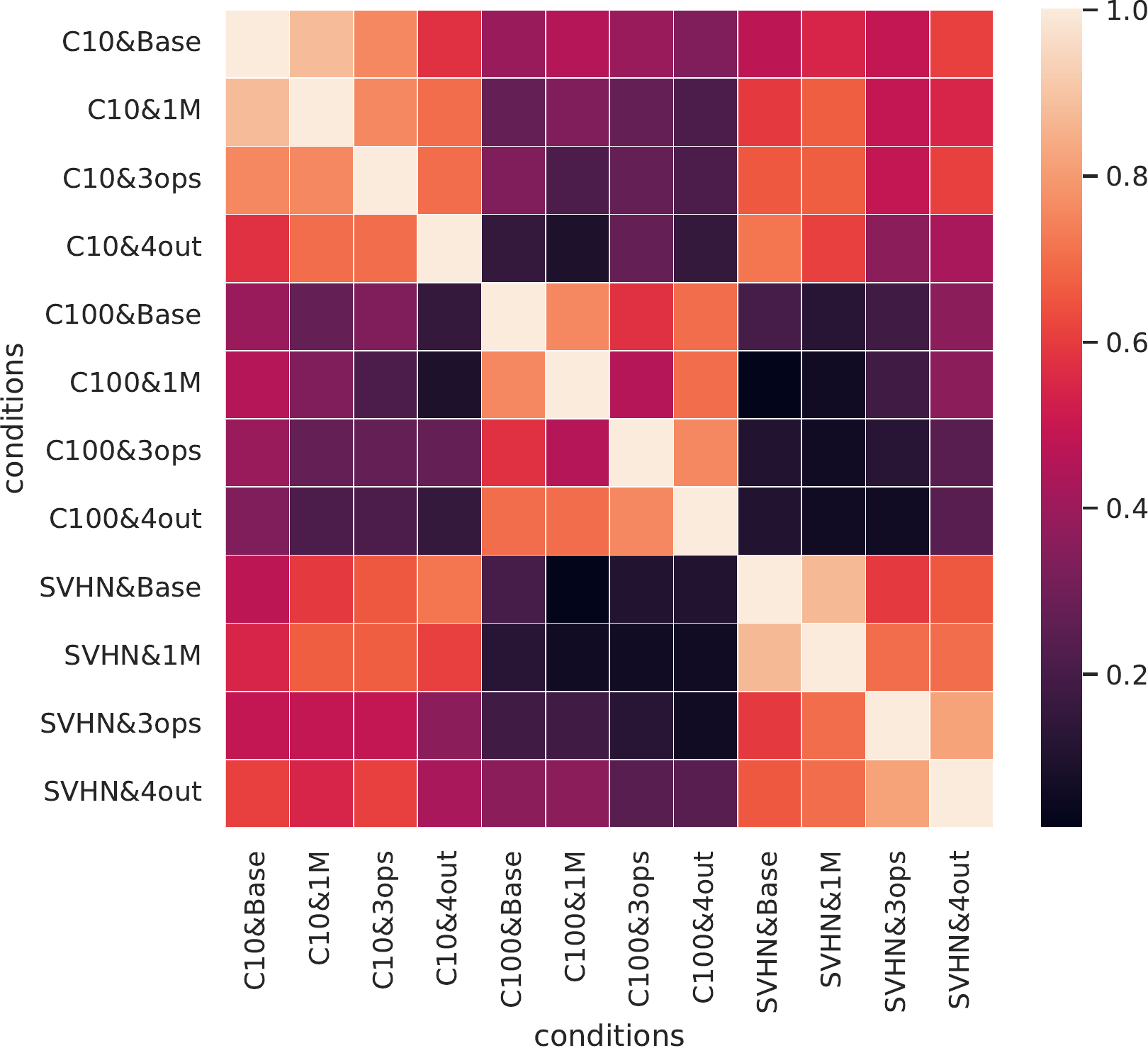} 
\caption{Small KD corresponds to low rank correlation. Method performs well on one condition does not guarantee its precedence on others. Single condition evaluation to claim superior could be misleading and not generalizable.}
\label{fig10}
\vskip -0.1in
\end{figure}

The main scores are reported in Table~\ref{table5}. Figure~\ref{fig8} illustrates the distribution of the search results in terms of depth, \textit{val\_acc} and \textit{\#param}. Figure~\ref{fig9} shows the results grouped by policies on the left and shows the performance differences between policies on the right. Kendall's Tau (KD $\tau$) is widely used to study the rank correlations on NAS~\cite{yu2019evaluating,park2020towards,zhang2020deeper}. Figure~\ref{fig10} illustrates a heatmap of KD to show a pairwise correlation of the twelve baselines' ranking over twelve conditions. We also demonstrate the improvements of discernibility in Table~\ref{table6}.

\textbf{Observations from the results} (1$\sim$5): 
\textbf{1}. If a method prefers a deep and large cell, such as GAEA-ERM and PC-DARTS, it is often more difficult to learn the proper gradient path that deteriorates performance while scaling up from Base to 1M; 
\textbf{2}. In contrast, if a method prefers simple and shallow architecture, such as SP-DARTS and GDAS, it is likely to fail to yield good performance on the conditions prefer deeper structures, e.g. on C100;  
\textbf{3}. It's hard to balance the preferences at the same time, e.g. SP-DARTS is one of the top performant art on both C10 and SVHN but is poor on C100. PC-DARTS is just the opposite that works well on C100. But this dilemma is our very intention in designing LHD and balancing the contradictory is also our expectation of the superior method;
\textbf{4}. Both transductive robustness and discretizations have a significant impact on methods ranking as illustrated in Figure~\ref{fig10}; 
\textbf{5}. We notice that even with the improved discernibility of the space, there are still local indiscernible in ranks, e.g. AMARs of the top-3 methods are $\leq 0.1$ under some conditions in Table~\ref{table6}. Multi-condition evaluations compensate this that if a method claims superior, it should prove that across most conditions (if not all) rather than upon a single condition with marginal score differences. We specify more observations on the characteristics of methods in Appx.H.

\begin{figure}[t]
\centering
\includegraphics[width=1\columnwidth]{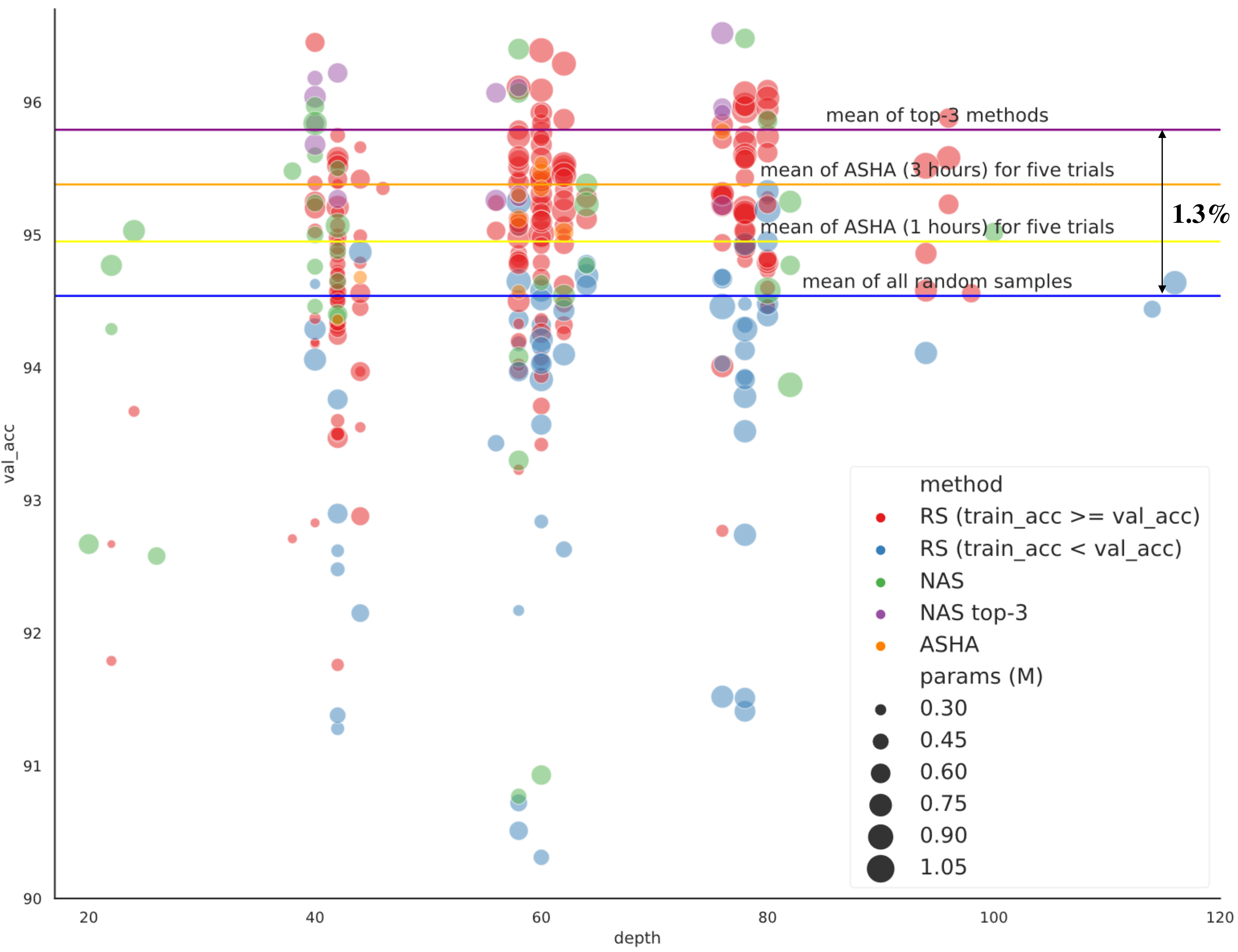} 
\caption{Random samples (RS) and random search (ASHA five trials) on C10\&Base. ASHA has been proven to be an art partial training method that outperforms leading adaptive search strategies for hyperparameter optimization~\cite{li2020system}. Results on C10\&3ops, C10\&4out and more details are provided in Appx.I.}
\label{fig13}
\vskip -0.1in
\end{figure}

\subsection{Random Sampling and Random Search}
We conduct further studies on LHD and observe (1$\sim$5):
\textbf{1}. \cite{yang2019evaluation} evaluated 200+ samples on DSS and observed ``all within a range of 1\% after a standard full training on C10" i.e. narrow  accuracy range. We evaluate 250+ random samples on C10\&Base which exhibit a much larger  accuracy gap in Figure \ref{fig13};
\textbf{2}. More than 25\% of the random samples have \textit{train\_acc}$<$\textit{val\_acc} after full 600 epochs of training which also occurs in some NAS methods that favor deep and large cell, both of which
imply an absence of appropriate gradient path and highlight the deliberate harder part of LHD.
\textbf{3}. The correlation between \textit{\#param} and \textit{val\_acc} is 0.29/0.26/0.20 for random samples on BASE/3ops/4out on C10 as opposed to 0.52 on C10\&DSS~\cite{ning2021evaluating}, demonstrating that LHD breaks the tight correlation~\cite{he2020milenas} under even lower \textit{\#param} regime. LHD thereby makes method more difficult to obtain high scores through overfitting \textit{\#param} and the validity of the architecture per se is more important;
\textbf{4}. As marked by Figure \ref{fig13}, the mean  accuracy of the top-3 methods outperforms random sampling by a large margin (1.3\%) on C10\&Base in contrast to many previous studies pointing to only a trivial gap ($<$0.5\%) between art NAS methods and random sampling on C10\&DSS~\cite{yu2020evaluating,yang2019evaluation,garg2020revisiting,lindauer2020best}, in particular underpins the discernable improvement in proposal. 
\textbf{5}. Random search is conducted by combining ASHA which is previously studied on DSS~\cite{li2020random} and shows competitive results. The NAS method only needs one-shot 3-hour training to obtain search results for all policies (Base/3ops/4out) whereas ASHA must be applied separately, so we provide both 1-hour and 3-hour results for ASHA in Figure~\ref{fig13}.

\begin{table}[!t]
\centering
\vskip -0.1in
\caption{Evaluate different settings. RA denotes RandAugment~\cite{cubuk2020randaugment} and CM denotes CutMix~\cite{yun2019cutmix}. $i$ controls the drop-path rate which is also positively correlated with the cells' interconnection density.}
\vskip 0.05in
\label{table7}
\resizebox{0.47\textwidth}{!}{
\begin{tabular}{c|c|ccc}
\toprule
\multirow{2}{*}{Dataset} & \multirow{2}{*}{Architecture} & \multicolumn{3}{c}{Evaluations} \\ \cline{3-5} 
 &  & \multicolumn{1}{c|}{\textit{\#param} (M)} & \multicolumn{1}{c|}{Settings} & \textit{val\_acc} (\%) \\ \hline
\multirow{3}{*}{C10} & SpC10-B & \multicolumn{1}{c|}{0.49} & \multicolumn{1}{c|}{$i$=0.04,  RA+CM} & 96.79 \\ \cline{2-5} 
 & \multirow{2}{*}{Rs3ops} & \multicolumn{1}{c|}{0.8/2.5/3.5} & \multicolumn{1}{c|}{$i$=0.06/0.07/0.08} & 96.84/97.51/97.70\\ \cline{3-5}
                      &                         &  \multicolumn{1}{c|}{2.5/3.5}     &  \multicolumn{1}{c|}{$i$=0.05/0.06, RA} &  97.69/97.82 \\
 \hline
SVHN & SpSVHN-B & \multicolumn{1}{c|}{0.58} & \multicolumn{1}{c|}{different recipe} & 96.64\\
\bottomrule
\end{tabular}
}
\vskip -0.1in
\end{table}

\begin{figure}[!t]
\centering
\includegraphics[width=1.0\columnwidth]{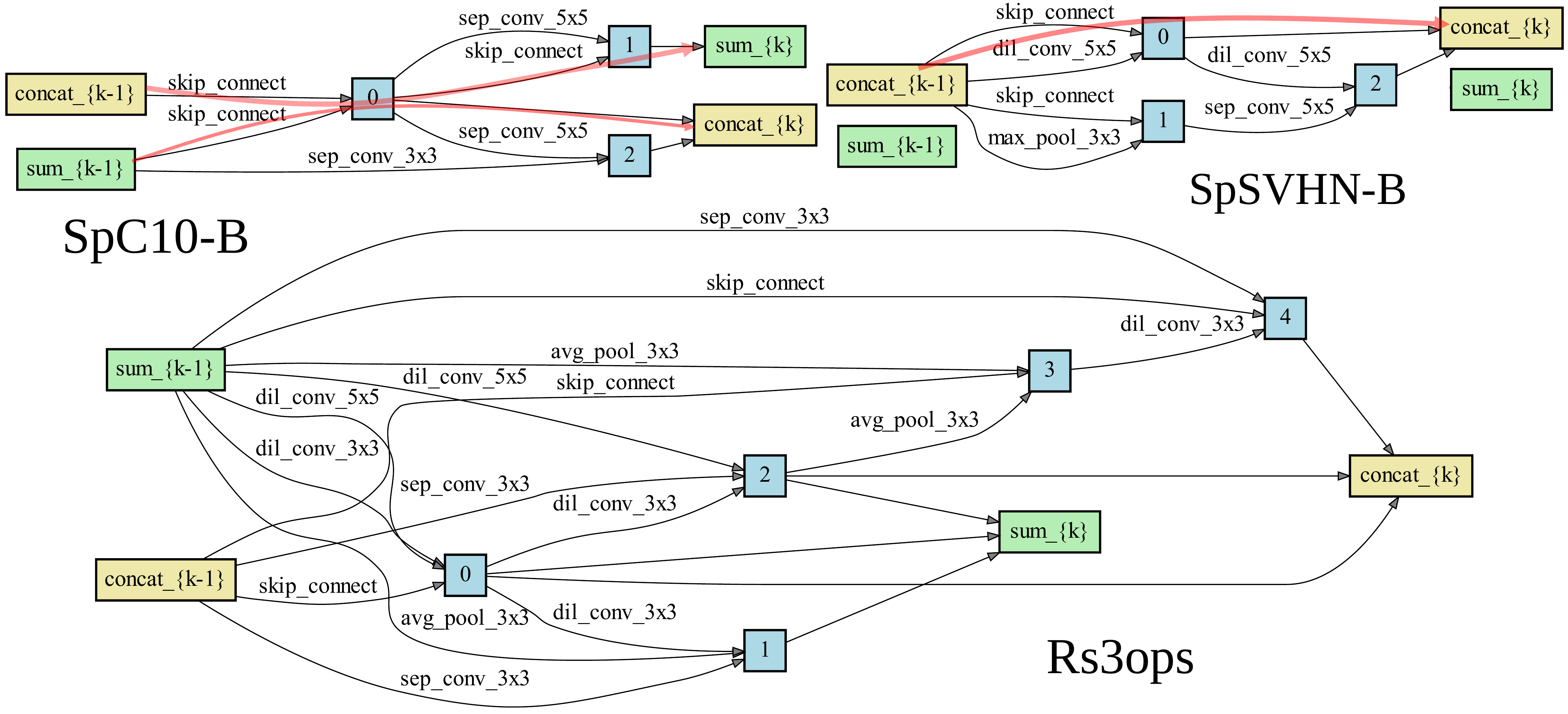} 
\vskip -0.1in
\caption{High performant normal cells with simple connection patterns searched by SP-DARTS on C10\&Base and SVHN\&Base, named SpC10-B and SpSVHN-B respectively. Learned gradient paths are evident and highlighted in red. The Rs3ops is much more complex that comes from random sampling on C10\&3ops. }
\label{fig11}
\vskip 0.1in
\includegraphics[width=0.8\columnwidth]{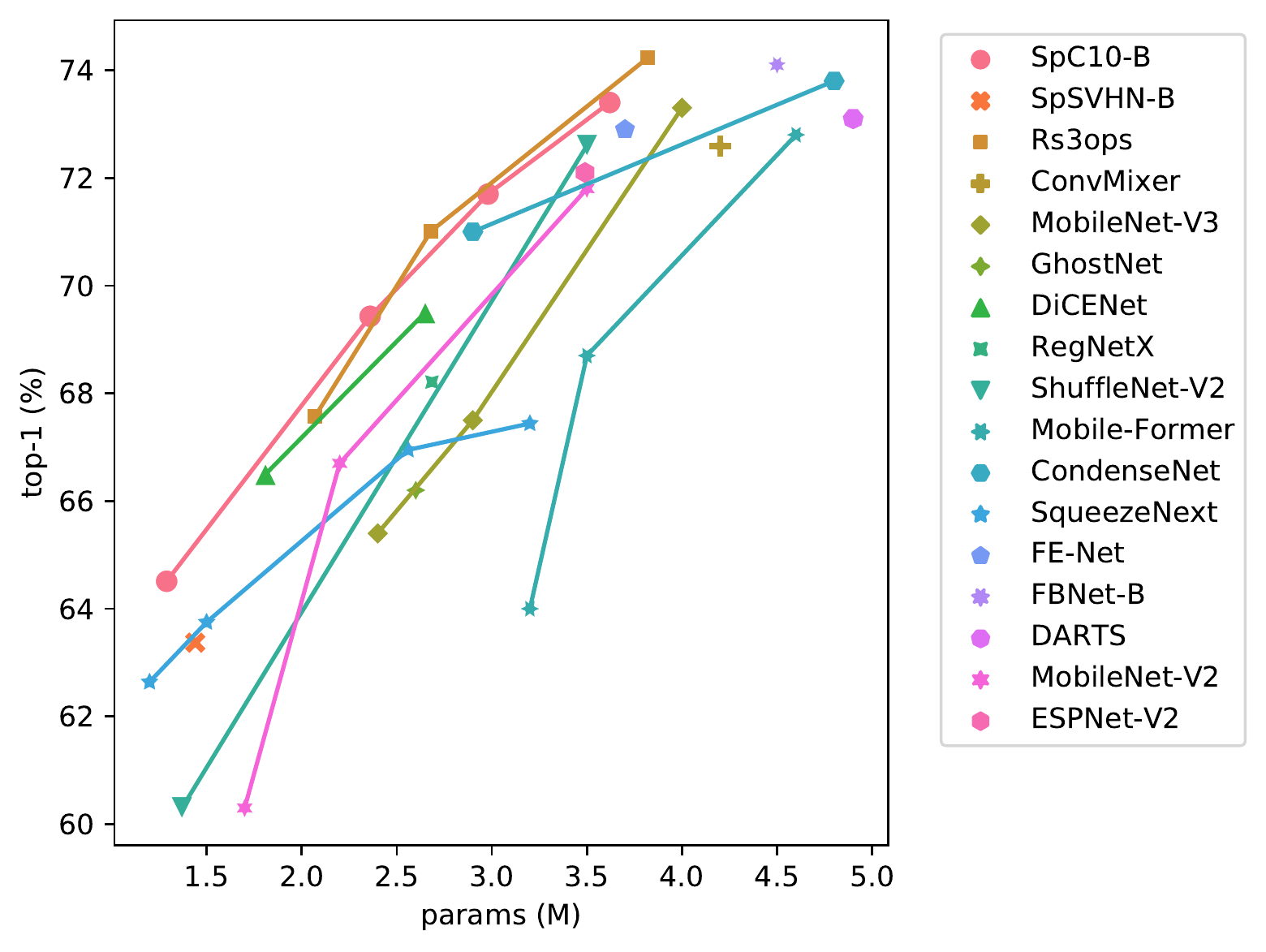} 
\vskip -0.1in
\caption{Comparisons on ImageNet-1k under mobile regime.}
\label{fig12}
\vskip -0.1in
\end{figure}

\section{Beyond the Benchmark}
We visualize three high performant normal cells in Figure~\ref{fig11}. Both SpC10-B and SpSVHN-B retain only three out of five intermediate nodes in the space due to removable search in proposal. Furthermore, method can obtain one input/output structure like SpSVHN-B that underscores the effectiveness of the s/e output blocks in finding inter-cell connection patterns. Additionally, We can easily recognize the learned gradient paths in Figure~\ref{fig11} that replace the fixed inter-cell skip connections in DSS. Beside the simple cells, LHD is also capable to encode rather complicated connection patterns as shown by Rs3ops on the bottom pane.

We try to ablate the evaluation settings and ultimately deliver a strong scores 96.79\% on C10 with only 0.49M \textit{\#param} in Table~\ref{table7}. For comparison, ConvMixer \cite{trockman2022patches} equipped with RA+CM+Mixup+Random Erasing attains the best 95.19\% (0.594M) and 95.88\% (0.707M) in their ablations. Besides, Rs3ops yields on par or superior \textit{val\_acc} with 30\% lesser \textit{\#param} comparing with the optimal results on DSS in Table~\ref{table1}. We also test another drastically different recipe where DenseNet40\_k12 delivers 96.22\% with similar \textit{\#param} but 2$\times$ more MACs under the same recipe on SVHN\footnote{longrootchen/densenet-svhn-classification-pytorch}.
To evaluate the transferability on ImageNet (224$\times$224), we closely follow the evaluation recipe on DSS that moderately increase the capacity to accommodate higher resolution inputs. Our scores are competitive against a wide range of baselines shown in Figure~\ref{fig12}.

\section{Compare with Queryable Benchmark}
Tabluar benchmark has an unquestionable efficiency upside. However, its downside also comes from the efficiency that makes methods easier to be tuned and overfit. We have witnessed a fair number of studies claimed to stably achieve the global optimal on NB201~\cite{dong2019bench}, but none of these methods yield that level of superior on DSS or LHD;  
The recent NAS-Bench (NB) study \cite{mehta2021bench} underscores the similar point in studying non-one-shot methods on more benchmarks. \citet{mehta2021bench} even appealed to stop focusing much on smaller NB201~\cite{dong2019bench} and NB101~\cite{zela2019bench,ying2019bench} and rather embrace larger and novel new NBs. One-shot methods can be applied on four NBs in \cite{mehta2021bench}  and three of which, including NB201 and NB101, are tabular benchmarks in addition to DSS. Since evaluating NAS only on tabular benchmarks is always considered inadequate, our work is one crucial complement rather than an exclusivity to the counterparts, in particular for one-shot methods and topology search.
On the other hand, tabular benchmarks hardly provide effective insight for other related fields due to their limited size of spaces. In contrast, more practical DSS has inspired follow-up researches in various forms~\cite{shu2019understanding,han2021learning,knyazev2021parameter}.
Surrogate benchmark~\cite{siems2020bench} is another type of queryable NB that predicts space statistics by pre-training tens of thousands of architectures. Surrogating twelve conditions in our case requires pre-training about one million samples which it's currently unrealistic.

\section*{Acknowledgements}
This work is supported by the National Key R \& D Program of China (2022YFF0503900)

\section{Conclusion}
In this paper, we dig into hardening and enlarging the canonical benchmark space DSS under limited resources, taking care of both discernibility and accessibility. We conduct a comparative study to establish a multi-condition evaluation benchmark and focus on comparing the unique contribution of each method but leaving their possible combinations for the future work. In particular, we provide abundant art baselines and all the scores can be used out of box without laborious repetition.
For fair comparison, we strongly recommend practitioners to only tune on one condition and transfer the exact settings to others. The results after tuned can be provided separately if necessary. We believe that the benefits of our study are multifaceted as we provide a basis for the further research, including a versatile and inclusive search space, a more revealing and all accessible benchmark and the research progress of the fair comparison of methods.  


\bibliography{icml23}
\bibliographystyle{icml2023}

\newpage
\appendix
\onecolumn
\section{Related Works}
As much of the related works were already mentioned in the introduction, we highlight several fields of the closely related works separately in this section.

\textbf{Topology search}: A handful of previous studies focused on the search space some of which pointed out that the search space is underdeveloped compared to the rapid progress of NAS methods. \citet{xie2019exploring} underscored that the success of many hand-designed networks comes from the innovation of the connection pattern. They demonstrated the network topology generated by different random strategies can clearly affect performance. \citet{shu2019understanding} made the point that the connection pattern rather than the operation selection significantly affects the landscape of the gradient and thereby affect the convergence speed of the network. They claimed that this observation can be used as a guideline for the network design in the future. Besides, one of the most prominent difference between hyperparameter optimization and NAS is that NAS can search connection topology among different operations and layers \cite{zoph2018learning,garg2020revisiting} which is also the key ingredient to increase the capacity of search space and find efficient networks \cite{real2020automl,wortsman2019discovering}.

\textbf{Single path (slimmable) search spaces}: As another line of the gradient-based methods came up with single-path MBConv-based search spaces that are usually built on searching a combination of channel numbers, input resolutions, network depths, expansion ratio (width multipliers) where the interconnectivity patterns between operations are largely constrained \cite{yu2020bignas,wang2021attentivenas,dong2021efficientbert,stamoulis2019single,peng2020cream,huang2021searching,wu2021fbnetv5,shen2022efficient}. Single-path NAS regularly involves two-stage decoupled optimization \cite{ren2021comprehensive} in which the training of supernet on ImageNet generally employ at least eight GPUs and costs hundreds of GPU hours at a minimum \cite{chen2021searching}. In contrast, DSS is mostly conducted by searching operation selections and their interconnection patterns simultaneously and costs only a few hours on a single GPU under low sample sizes. In summary, DSS is important for NAS, not only for the search of the connection topology, but also for the accessible benchmark. Still, how to design a more general, flexible and free of human bias search space will remain challenging and advantageous for the NAS community for a long time \cite{he2021automl}.

\textbf{Benchmark on DSS}: We notice some of the most recent studies of benchmark dedicated to the evaluation of multitask and transferability. 
These studies are often limited by the size of the search space~\cite{duan2021transnas}, or use non-standard benchmark datasets resulting in few available baselines~\cite{tu2021bench}. Our benchmark is evaluated on the most commonly used standard benchmark datasets (CIFAR-10, CIFAR-100, SVHN) so that it is easy to find a large number of available baselines (handcrafted and NAS) within the latest literatures. 
Another irreplaceability of the DSS is that there are already extensive methods that are finetuned, provide their implementations or report their scores on DSS~\cite{he2020milenas,xu2019pc,li2021geometry,chen2021drnas,xue2021rethinking,zhang2021robustifying,wang2020rethinking,chu2020darts,dong2019searching,chen2020stabilizing,chu2020fair,arber2020understanding,lee2020rapid}, which can be used directly as the competitive baselines for DSS-based evaluations without requiring the researchers to re-implement or even re-execute the experiments. If the scores are not available out of the box, expecting researchers to reimplement multiple art baselines on a new codebase is often labor-intensive or even unachievable that also tend to get caught up in unfair comparison controversy without adequate tuning in this case. For these reasons, our study follows the configuration of the DSS benchmark to the greatest extent possible (tasks, datasets, search and evaluation protocols) so that our claim of transductive robustness is sufficiently convincing. Second and most importantly, we provide the evaluation scores of established baselines~\cite{liu2018darts,he2020milenas,chen2021drnas,li2021geometry,dong2019searching,zhang2021robustifying,xu2019pc,chu2020darts,luo2022surgenas,ye2022b} so that the researches can use the benchmark almost out of the box by only implementing their own methods and compare with the scores we provided in the main text . \textit{Our work is not to construct a new search space and try every existing method to see which can bring us the good arch to challenge the architecture. On the contrary, our work actually embarrasses existing methods by removing their dependent artifacts, enlarging search space, searching upon different reasonable conditions}. Similar to our work, \citet{arber2020understanding} pioneered to tailor DSS into four search spaces S1$\sim$S4 which are specially customized to embarrass DARTS and widely used to validate the robustness of the regularization methods. 

\textbf{Tabular and surrogate benchmarks}: NAS-BENCH-101 \cite{ying2019bench} is inappropriate as our counterpart since NB101 cannot be used by one-shot NAS methods including DARTS and its variants. NAS-BENCH-1Shot1 \cite{zela2019bench} made this issue very clear and proposed to map NB101 to three available search spaces. Nevertheless, NB1S1 was still rarely used for evaluation by current NAS methods, especially compared to its concurrent tabular work NAS-BENCH-201 \cite{dong2019bench}, due to some internal limitations e.g. i) few available ops (only three); ii) no explicit methods ranking in the paper main text; iii) unfriendly to implement new methods. Besides, NB101 (NB1S1) is a tabular benchmark and the evaluation on tabular benchmarks is never enough for NAS methods due to the highly limited search space ($10^5$ compared to $10^{58}$ of DSS and $10^{84}$ of our LHD). Furthermore, tabular benchmarks hardly provide effective insight for the practical network design. In contrast, we have seen that more realistic DSS has inspired several studies \cite{shu2019understanding,knyazev2021parameter}. NAS-BENCH-301~\cite{siems2020bench} adopts a surrogate-based methodology on DSS that predicts performances with the performances of about 60k anchor architectures.

\begin{table}[!t]
\caption{Softmax attends on edge in DSS as opposed to it attends on aggregation within \textbf{input end} of nodes in LHD i.e. the origin of \textbf{input}-softmax.}
\label{table2}
\vskip 0.1in
\resizebox{\textwidth}{!}{
\begin{tabular}{ccccc}
\toprule
Search space & Inputs of s/e cell ${S^t}$& 
Outputs of ${S^t}$ and ${U^t}$& Node ${x_j}$ aggregation & Edge ${g_{i,j}}({x_i})$ operation \\
\midrule 
DSS & $F_U^{t - 1}$, $F_U^{t - 2}$ & \begin{tabular}[c]{@{}c@{}}$F_S^t = \left\{ {x_j^t|j \in ({n_i}..N]} \right\}$ \\ where ${n_i} = 2$, $F_U^t = M_ \oplus ^{1 \times 1}\left( {F_S^t} \right)$ \end{tabular} & ${x_j} = \sum\nolimits_{i < j} {{g_{i,j}}({x_i})}$  & \begin{tabular}[c]{@{}c@{}}${g_{i,j}}({x_i}) =  \left\langle {{\bf{a}}_{i,j}},{O_{i,j}}({x_i}) \right\rangle $ \\ where ${{\bf{a}}_{i,j}} = {\rm{softmax}} ({A_{i,j}})$ \end{tabular} \\
\midrule 
LHD & $F_ + ^{t - 1}$, $F_ \oplus ^{t - 1}$ & \begin{tabular}[c]{@{}c@{}}$F_ + ^t =  \left\langle{\bf{p}}_ + ^t,{{\bf{x}}^t} \right\rangle $ where \\ ${\bf{p}}_ + ^t = {\rm{softmax}} (P_ + ^t)$, $P_ + ^t = \left\{ {\left[ {{\beta _ + }} \right]_j^t|j \in ({n_i}..N]} \right\}$ \\ $F_ \oplus ^t = M_ \oplus ^{1 \times 1}\left( {{\bf{p}}_ \oplus ^t \circ {{\bf{x}}^t}} \right)$ where \\ ${\bf{p}}_ \oplus ^t = \left\{ {{\rm{sigmoid}}(\left[ {{\beta _ \oplus }} \right]_j^t)|j \in ({n_i}..N]} \right\}$, \\ ${{\bf{x}}^t} = \left\{ {x_j^t|j \in ({n_i}..N]} \right\}$ and ${n_i} = 2$ \end{tabular} & \begin{tabular}[c]{@{}c@{}}${x_j} = \left\langle {{{\bf{a}}_j},{g_j}} \right\rangle $ where \\ ${{\bf{a}}_j} = {\rm{softmax}} ({A_j})$, \\ ${A_j} = \left\{ {\alpha _{i,j}^m|i \in [1..j),m \in [1..M]} \right\}$, \\ ${g_j} = \{ {g_{i,j}}({x_i})|i \in [1..j)\} $ \end{tabular} & \begin{tabular}[c]{@{}c@{}} ${g_{i,j}}({x_i}) = {O_{i,j}}({x_i})$ \\ $= \{ {o^m}\left( {{x_i}} \right)|m \in [1..M]\}$ \end{tabular} \\
\bottomrule
\end{tabular}
}
\caption{Hyperparameter settings of baselines in search. DARTS settings are closely follow the released code on DSS. Specific settings of other baselines follow their released code on DSS as well. All settings keep consistent across C10, C100, SVHN.}
\vskip 0.1in
\label{table9}
\resizebox{\textwidth}{!}{
\begin{tabular}{lccccccccccl}
\toprule
method & batch\_size & learning\_rate & learning\_rate\_min & momentum & weight\_decay & epochs & init\_channels & layers & arch\_learning\_rate & arch\_weight\_decay & additional \\
\midrule 
DARTS & 176 & 0.025 & 3e-4 & 0.9 & 3e-4 & 50 & 16 & 8 & 3e-4 & 0 & - \\
\midrule 
MiLeNAS & - & - & - & - & - & - & - & - & - & - & $\lambda=1$ \\
\midrule 
DrNAS & - & - & - & - & - & - & - & - & - & - & - \\
\midrule 
GAEA-B & - & - & - & - & - & - & - & - & 0.1 & - & - \\
\midrule 
GAEA-E & - & - & - & - & - & - & - & - & 0.1 & - & - \\
\midrule 
GDAS & 256 & 0.05 & - & - & - & - & - & - & - & - & \begin{tabular}[l]{@{}l@{}}tau\_min=0.1,\\ tau\_max=10\end{tabular} \\
\midrule 
SP-DARTS & - & - & 0.025 & - & - & - & - & - & - & - & \begin{tabular}[l]{@{}l@{}}warmup=5,\\ temp=0.0015\end{tabular} \\
\midrule 
PC-DARTS & 576 & 0.1 & - & - & - & - & - & - & 6e-4 &  & \begin{tabular}[l]{@{}l@{}}warmup=15,\\ k=4\end{tabular}\\
\midrule 
DARTS- & - & - & - & - & - & - & - & - & - &  & \begin{tabular}[l]{@{}l@{}}$\beta$=1\\decay scheme\\:linear ($\beta\to$0)\end{tabular}\\
\midrule 
$\beta$-DARTS & - & - & - & - & - & - & - & - & - &  & \begin{tabular}[l]{@{}l@{}}weight scheme\\ =0$\to$100\end{tabular}\\
\midrule 
SurgeNAS & - & - & - & - & - & - & - & - & - &  & \begin{tabular}[l]{@{}l@{}}$\beta$=1\\decay scheme\\:linear ($\beta\to$0)\end{tabular}\\
\bottomrule
\end{tabular}
}
\caption{Hyperparameter settings of evaluation. All baselines strictly follow the same.}
\vskip 0.1in
\label{table10}
\resizebox{\textwidth}{!}{
\begin{tabular}{ccccccccc}
\toprule
batch\_size & learning\_rate & learning\_rate\_min & momentum & weight\_decay & epochs & init\_channels & stacked cells & data augmentations (cutout, flip, crop) \\
\midrule 
64 & 0.025 & 3e-4 & 0.9 & 3e-4 & 600 & \begin{tabular}[c]{@{}c@{}}36 for Base,\\ 3ops and 4out \end{tabular}& \begin{tabular}[c]{@{}c@{}}25 for 1M,\\ 20 for others\end{tabular} & true for C10 and C100, false for SVHN\\
\bottomrule
\end{tabular}
}
\end{table}

\section{Formulations of LHD}
We formalize LHD in Tabel~\ref{table2} where $\beta$ is the path parameter and $ \circ $ denotes Hadamard product. $F$ represents the cell outputs and subscripts $U$ and $S$ refers to s/e and u/e. Subscripts $ + $ and $ \oplus $ represent summation and concatenation. We omit the cell index $t$ within node and edge columns and don’t distinguish coessential set and vector for brevity.

\section{Complexity of the Continuous DAGs}
In LHD, cell accommodates five intermediate nodes with 2+3+4+5+6=20 inter-connection compound edges each of which factors in $7$ operations thus a total of ${2^7}$ combinations. For a single CSB cell output, the substructure complexity of the continuous space is ${2^{7 \times 20}} = {2^{140}} \approx {10^{42}}$ without considering graph isomorphism. Two searchable cells, normal and reduction, account for the total complexity at least ${10^{42 \times 2}} = {10^{84}}$ for LHD. Softmax in DSS is applied on edge implying at least one operation will be selected.  We subtract the case where no operation is selected on each edge and get the total complexity of DSS as ${\left( {{2^7} - 1} \right)^{14 \times 2}} \approx {10^{58}}$. We have to note that the post-search discretization will introduce a large amount of inductive bias and determine the valid subspace smaller than the total capacity of the continuous DAG.

\begin{table}[t]
\caption{Evaluation results on CIFAR-100 and SVHN on LHD.}
\vskip 0.1in
\label{table8}
\resizebox{\textwidth}{!}{
\begin{tabular}{l|ccc|ccc|ccc|ccc}
\toprule
C100           & \multicolumn{3}{c|}{Base}            & \multicolumn{3}{c|}{1M}              & \multicolumn{3}{c|}{3ops}           & \multicolumn{3}{c}{4out}            \\ 
Method         & \textit{val\_acc} (\%)   & \textit{\#param} & top-1/top3  & \textit{val\_acc} (\%)   & \textit{\#param} & top-1/top3  & \textit{val\_acc} (\%)  & \textit{\#param} & top-1/top3  & \textit{val\_acc} (\%)  & \textit{\#param} & top-1/top3  \\
\midrule
DARTS          & 69.18±3.85  & 0.62     & 72.68/71.97 & 70.38±4.37  & 1.55     & 74.17/73.54 & 70.27±2.51 & 0.71     & 73.18/72.01 & 70.82±2.20 & 0.9      & 72.87/72.41 \\
DrNAS          & 53.92±14.67 & 0.5      & 66.90/64.16 & 49.65±18.55 & 1.54     & 68.89/62.56 & 61.69±3.59 & 0.53     & 63.77/63.38 & 47.21±6.21 & 0.85     & 54.69/51.35 \\
GAEA-B         & 71.51±3.64  & 0.63     & 75.28/73.47 & 73.59±2.93  & 1.53     & 76.71/75.14 & 73.58±1.67 & 0.69     & 75.57/74.43 & 74.14±1.52 & 1        & 76.23/75.06 \\
GAEA-E         & 67.81±1.03  & 0.86     & 69.12/68.47 & 67.06±1.95  & 1.54     & 69.90/68.26 & 71.07±0.40 & 1.06     & 71.51/71.35 & 68.10±1.13 & 1.21     & 69.34/68.86 \\
GDAS           & 62.99±1.60  & 0.51     & 64.26/64.15 & 65.97±0.91  & 1.53     & 66.85/66.63 & 61.89±4.31 & 0.58     & 66.45/64.36 & 65.07±2.04 & 0.98     & 67.31/66.44 \\
MiLeNAS        & 68.98±0.97  & 0.65     & 70.00/69.61 & 70.32±0.98  & 1.57     & 71.25/70.88 & 69.17±1.79 & 0.74     & 70.58/70.35 & 69.77±0.47 & 0.91     & 70.48/70.02 \\
PC-DARTS       & 71.03±5.32  & 0.77     & 74.04/73.73 & 71.52±5.64  & 1.54     & 74.86/74.75 & 71.78±5.18 & 0.9      & 74.95/74.38 & 72.12±4.63 & 1.05     & 75.65/74.91 \\
Random         & 69.14±2.79  & 0.7      & 72.08/70.93 & 69.34±5.78  & 1.54     & 74.46/73.13 & 70.66±2.24 & 0.79     & 73.18/72.17 & 70.71±4.17 & 0.97     & 74.84/73.64 \\
SP-DARTS       & 68.99±1.13  & 0.52     & 70.69/69.60 & 70.97±0.71  & 1.55     & 71.89/71.44 & 66.88±0.92 & 0.55     & 68.08/67.47 & 69.15±1.60 & 0.84     & 70.90/70.06 \\
DARTS-         & 68.96±2.55  & 0.65     & 71.44/70.75 & 67.82±3.67  & 1.54     & 72.53/70.20 & 61.26±6.03 & 0.74     & 67.82/64.58 & 57.67±6.85 & 0.9      & 65.41/62.11 \\
$\beta$--DARTS & 70.03±4.28  & 0.62     & 73.60/72.25 & 72.35±3.68  & 1.54     & 75.52/74.45 & 71.35±2.28 & 0.68     & 73.44/72.97 & 72.33±2.31 & 0.93     & 74.90/73.96 \\
SurgeNAS       & 69.42±3.59  & 0.88     & 73.00/71.26 & 69.88±3.94  & 1.56     & 74.04/72.01 & 69.74±1.78 & 1.01     & 71.02/70.83 & 69.78±2.52 & 1.11     & 72.91/71.20\\
\midrule
\midrule
SVHN           & \multicolumn{3}{c|}{Base}           & \multicolumn{3}{c|}{1M}             & \multicolumn{3}{c|}{3ops}           & \multicolumn{3}{c}{4out}            \\ 
Method         & \textit{val\_acc} (\%)  & \textit{\#param} & top-1/top3  & \textit{val\_acc} (\%)  & \textit{\#param} & top-1/top3  & \textit{val\_acc} (\%)  & \textit{\#param} & top-1/top3  & \textit{val\_acc} (\%)  & \textit{\#param} & top-1/top3  \\
\midrule
DARTS          & 95.77±0.70 & 0.58     & 96.36/96.26 & 95.94±0.59 & 1.55     & 96.52/96.35 & 96.00±0.50 & 0.68     & 96.43/96.34 & 95.58±0.58 & 0.87     & 96.20/95.96 \\
DrNAS          & 96.58±0.26 & 0.67     & 96.81/96.77 & 96.65±0.18 & 1.55     & 96.83/96.78 & 96.66±0.12 & 0.81     & 96.78/96.75 & 96.61±0.10 & 1.09     & 96.71/96.68 \\
GAEA-B         & 96.85±0.10 & 0.54     & 96.98/96.90 & 97.01±0.10 & 1.57     & 97.17/97.07 & 96.78±0.06 & 0.68     & 96.84/96.83 & 97.06±0.08 & 0.98     & 97.19/97.11 \\
GAEA-E         & 96.68±0.07 & 0.87     & 96.76/96.73 & 96.63±0.02 & 1.54     & 96.66/96.64 & 96.58±0.07 & 1.06     & 96.67/96.64 & 96.63±0.06 & 1.05     & 96.68/96.67 \\
GDAS           & 96.81±0.17 & 0.8      & 97.07/96.92 & 96.95±0.10 & 1.54     & 97.08/97.03 & 96.67±0.46 & 0.91     & 97.10/96.92 & 96.62±0.48 & 1.02     & 97.21/96.95 \\
MiLeNAS        & 96.71±0.20 & 0.51     & 96.92/96.83 & 96.89±0.18 & 1.57     & 97.03/97.00 & 96.67±0.22 & 0.67     & 96.85/96.83 & 96.76±0.19 & 1        & 97.00/96.88 \\
PC-DARTS       & 96.69±0.08 & 0.82     & 96.82/96.73 & 96.87±0.13 & 1.53     & 97.03/96.94 & 96.68±0.17 & 0.97     & 96.91/96.80 & 96.82±0.07 & 1.08     & 96.88/96.86 \\
Random         & 96.81±0.24 & 0.64     & 97.09/96.97 & 96.93±0.20 & 1.52     & 97.16/97.07 & 96.54±0.62 & 0.73     & 96.99/96.92 & 96.66±0.35 & 0.96     & 97.05/96.89 \\
SP-DARTS       & 96.78±0.16 & 0.55     & 97.04/96.88 & 97.00±0.21 & 1.57     & 97.29/97.12 & 96.93±0.12 & 0.71     & 97.12/96.99 & 96.90±0.15 & 0.96     & 97.04/97.00 \\
DARTS-         & 94.45±0.59 & 0.54     & 95.17/94.86 & 94.66±0.47 & 1.56     & 95.15/94.96 & 95.23±0.24 & 0.63     & 95.59/95.37 & 94.70±0.46 & 0.85     & 95.17/94.96 \\
$\beta$--DARTS & 93.87±1.92 & 0.52     & 95.90/93.87 & 94.06±2.07 & 1.56     & 95.94/94.06 & 94.78±1.10 & 0.58     & 95.73/94.78 & 93.65±1.21 & 0.78     & 94.98/93.65 \\
SurgeNAS       & 96.86±0.11 & 0.84     & 96.96/96.89 & 96.90±0.06 & 1.57     & 97.01/96.94 & 96.77±0.06 & 0.93     & 96.84/96.82 & 96.83±0.08 & 1.05     & 96.98/96.87\\
\bottomrule
\end{tabular}
}
\end{table}
\begin{figure}[t]
\centerline{\includegraphics[width=\textwidth]{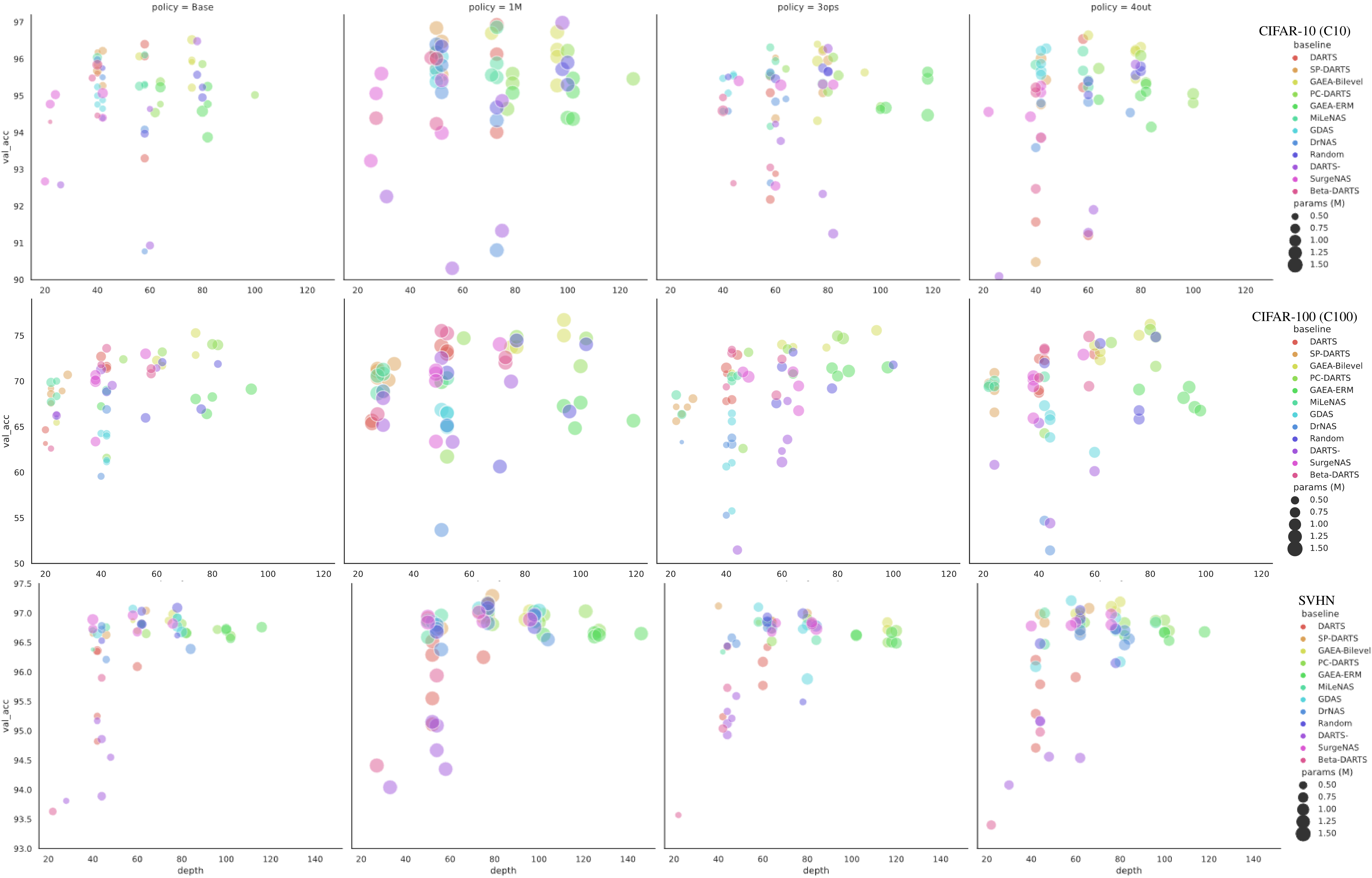}}
\caption{Distributions of the search results on \textit{val\_acc} versus depth coordinate frame. Depth refers to the number of sequential convolution layers within the longest path without counting stem. Depth has little effect on C10 performance, but deeper networks tend to achieve better C100 performance, whereas upper bound of the performance on SVHN is some kind negatively correlated with depth.}
\label{fig8_a}
\end{figure}

\begin{figure}[t]
\centering
\includegraphics[width=\textwidth]{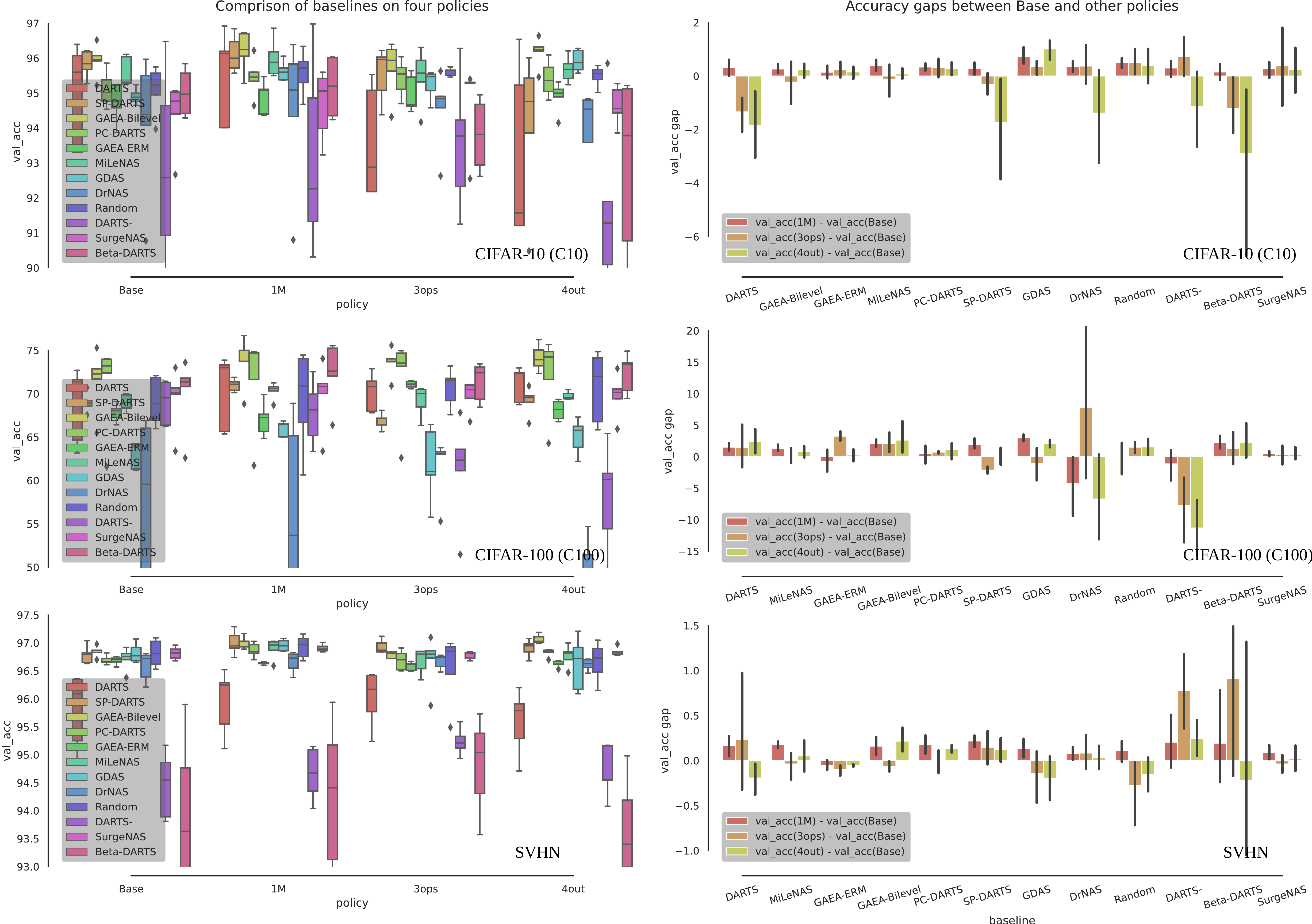} 
\caption{Baseline performances grouped by policies (left); \textit{val\_acc} gaps between Base and other policies (right).}.
\label{fig9_a}
\end{figure}

\section{Baseline Settings in Search and Evaluation}
Hyperparameter settings in search across baselines and the consistency settings of evaluation are provided in Table~\ref{table9} and Table~\ref{table10} respectively. Same settings are strictly followed on all conditions.

We provide the full source code in the repository 
and we also report some implementation details here for self-contained. Most of the implementation of baselines are search space agnostic, so our implementations are overall closely follow the released code from their authors listed at the last column of Table 3 in the main text.

For DARTS on LHD, we closely follow the source code released on DSS. The only special clarification needed is that the parameters of the output path of both CSB and SSB are initialized sampling from $N(0,1)$ and scaling the sample by $e$-$3$ which is in line with the initialization of the architecture parameters that representing the significance of operations in DARTS.
The implementation of MiLeNAS is directly based on DARTS, except that the architecture parameters are updated by the cumulative gradients on the both training and validation sets rather than the validation set alone in DARTS.
GAEA-Bilevel modifies the general gradient of DARTS to the exponential version,
GAEA-ERM goes one step further and trains architecture parameters and operation weights simultaneously without splitting a validation set from training set. GDAS replaces torch.softmax with gumble-softmax in DARTS. On the other hand, DrNAS replaces torch.softmax with torch.dirichlet. We directly use the scheduler of the temperature coefficient of softmax in SP-DARTS on the LHD. We also follow the channel sampling implementation in DSS of  PC-DARTS and migrate it directly to the LHD.
For random sampling, we discretize the randomly sampled parameters of the operation significance and the output paths of CSB and SSB are selected randomly and independently according to the Bernoulli distribution. If the obtained network is invalid (open circuit), repeat the sampling process until a valid architecture is obtained. We use the linear decay from 1 to 0 which is the default setting in DARTS- for the shortcut of each edge within DARTS- and for the individually shortcut of each operation within SurgeNAS. The weight scheme of $\beta$-DARTS (0$\to$100) is also consistent with the released code on DSS and NB201.

\section{Additional Results of the Benchmark}

We report mean and std of \textit{val\_acc}s in Table~\ref{table8} as the main scores. We also report the average size of finalnets to uncover the preference of baselines in the perspective of parameter scale. We observe that some methods are trapped in rare failure cases (see DARTS and DrNAS on C10, GAEA-Bilevel and PC-DARTS on C100), so we report additional top-1 and top-3 scores in Table~\ref{table8}. Figure~\ref{fig8_a} illustrates the distribution of the search results in terms of depth, \textit{val\_acc} and parameter scale under the twelve conditions that the column and row of the picture group correspond to the policies and datasets respectively. The first column of Figure~\ref{fig9_a} compares baseline \textit{val\_acc}s on each policy and the second column exhibits the differences in \textit{val\_acc} between Base and other policies for each baseline to illustrate the effect of the policy on different methods. 

Software version for search and evaluation of the benchmark: torch 1.9, cuda 11.1, cudnn 8.2, driver version 460.67. But we also test the search and evaluation codes and verify the empirical memory overhead on more recent version: torch 1.10, cuda 11.3, cudnn 8.3 and driver 495.44. The total number of evaluated finalnets is 540 and the footprint of both search and evaluation is about 500 GPU days.

\section{Path Tuning (PT) based Result Selection}
\citet{wang2020rethinking} proposed a new finalnet selection method based on a separated PT phase after searching to replace parameter-value-based one-off pruning. We carefully investigate the paper as well as released code\footnote{https://github.com/ruocwang/darts-pt} and have the following findings:

\textbf{1}. PT needs to mask and evaluate operations one-by-one in the PT phase. The idea of PT is largely NAS method agnostic but highly specific to the search space and entangled to the space design;

\textbf{2}. Owe to ``method agnostic", we recognize that the PT can be apply to all the baselines in our benchmark but inevitably incurs non-trivial additional time overhead (See ``4"), thus unfair to compare with parameter-value-based one-off pruning selection (See ``5");

\textbf{3}. Due to ``space entangled", it's non-trivial to determine many implementation details because of the difference between DSS and LHD. For example, if we need to tune the cell output, or just the operation selections? If the output path needs to be tuned, whether the tuning is done jointly with the operation selections, or separately?

\textbf{4}. PT needs to mask operations on each edge in forward pass to calculate the \textit{val\_acc} loss, so its computational cost is closely related to the number of operations, nodes and edges in the search space. When the space is enlarged, the computational cost will also increase linearly;

\textbf{5}. Apart from ``4",PT has to tune supernet and select result individually for each valid space (Base, 3ops, 4out) like random search, but one-time pruning only needs to search once without any extra-search time overhead. Therefore, the overhead of PT will even exceed the search phase when it is applied to three valid spaces respectively.

For above reasons, we eliminate PT in current benchmark and leave its verification to the future work.

\begin{figure}[t]
\centerline{\includegraphics[width=0.49\textwidth]{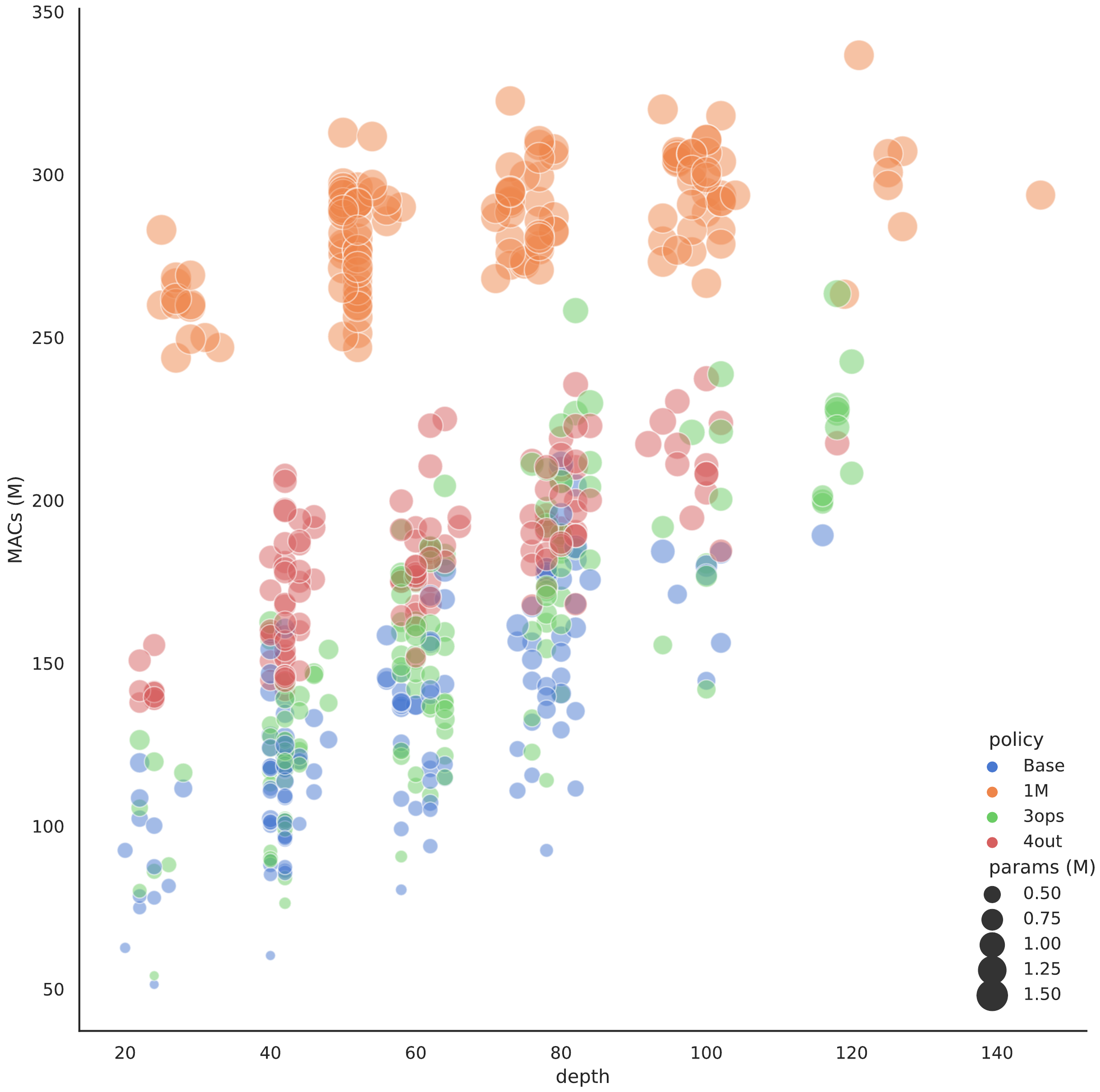}}
\caption{Distributions of the search results on MACs versus depth coordinates. The size of finalnets span a wide range of gap with a maximum 7$\times$ in terms of MACs, depth and parameters. By comparison, the depth and size are typically no more than 1.5$\times$ and 0.5$\times$ differences respectively of the search results on DSS.}
\label{fig14}
\end{figure}

\section{Heuristic Regularization in Evaluation}
DSS tightly couples search results and search space gives rise to that the gaps of depth and parameter scale of the finalnets released with source code never exceed 1.5$\times$ and 0.5$\times$ respectively. They elaborated a single recipe to evaluate all results on C10. In contrast, we observe a large variety of the search results in the lens of depth, flops, parameter scale due to the removable intermediate node of LHD and multiple discretization policies as shown in Figure~\ref{fig14} of our benchmark. This opens a new question did not appear on DSS, how to fairly evaluate search results with large differences in architecture.

Elaborating evaluation recipe for each finalnet is not our goal and quickly becomes intractable for a comprehensive evaluation. We aim to obtain reasonable and inter-comparable scores of the diverse results in the evaluation phase. Empirically, we observe that the regularization strength is the paramount factor affecting the performance of diverse architectures.

Similar to \cite{yang2019evaluation,arber2020understanding} we choose to overall closely follow the evaluation recipe of DSS across different datasets similar to previous practice in addition to which we propose a (tunable and adaptive) simple protocol to adjust the intensity of regularization heuristically for various conditions.
The regularization in evaluation recipe of DSS mainly involves data augmentation (Crop, flip, cutout) and drop path.

For evaluation phase of our benchmark, we adopt the same data augmentation on C10 and C100 and exclude it on SVHN. We come up a protocol ${r_{\rm DP}} = ic$ to adapt the drop path rate ${r_{\rm DP}}$ under different conditions where $c$ is the number of connections between intermediate nodes and concatenation output in the finalcell. $i$ is a tunable parameter across datasets and discrete policies. We first set $i$ as $0.01$ for Base, 3ops and 4out on C10 and increase it by 50\% for 1M due to the larger finalnet capacity. We double the value of $i$ on C100 and SVHN due to fewer samples per class and the exclusion of data augmentation respectively which make them both more likely to be overfitted.

\section{Observations of Methodological Characteristics from the Results of Benchmark}
Based on the benchmark results, we can make the following observations of the methods:

\textbf{1}. Search results of many baselines show clustering in the depth versus \textit{val\_acc} coordinates indicating the fixed preferences of the different methods in Figure~\ref{fig8_a}. 

\textbf{2}. We can actually get rich observations for each baseline from the left column of Figure~\ref{fig9_a}. For example, GAEA-ERM shows stable performance over different seeds, MiLeNAS non-trivially reduces the \textit{val\_acc} variance on 4out, PC-DARTS is policy-insensitive on C100, DARTS is more susceptible to the initialization seeds and always have greater performance fluctuations under all conditions compared to most baselines; 

\textbf{3}. Right column of Figure~\ref{fig9_a} shows that the baselines perform diversely on different policies. For example on C10, 4out severely deteriorates a number of baselines. GDAS, by contrast, shows remarkably superior scores on 4out than that on Base. Similarly, DrNAS and GAEA-ERM prefer 3ops but perform quite different on 4out;

\textbf{4}. As shown by Figure~\ref{fig8_a} and Table~\ref{table8}, both GAEA-ERM and PC-DARTS prefer larger and deeper cell while GDAS and SP-DARTS are just the opposite. For example on C10\&Base, the average parameter scale of GAEA-ERM is 65\% larger than that of GDAS, but the performance of GAEA-ERM is worse which highlights the challenging part of LHD that the methods are requisite to learn the appropriate gradient pathways autonomously rather than depending on hand-crafted skip connection;

\textbf{5}. SP-DARTS is one of the most performant methods on both C10 and SVHN but is poor on C100. PC-DARTS is just the opposite that performs well on C100. Failure cases are not uncommon among baselines. Both observations underpin the necessity to validate the search robustness across multiple datasets.

\textbf{6}. Silevel optimization is effective on stabilizing the training process shown by both GAEA-ERM and SurgeNAS.  Additionally, optimizing on mixlevel can be seen as an meaningful regularization to perform consistent across different discretization policies for DARTS.

\begin{figure}[t]
\centerline{\includegraphics[width=1\textwidth]{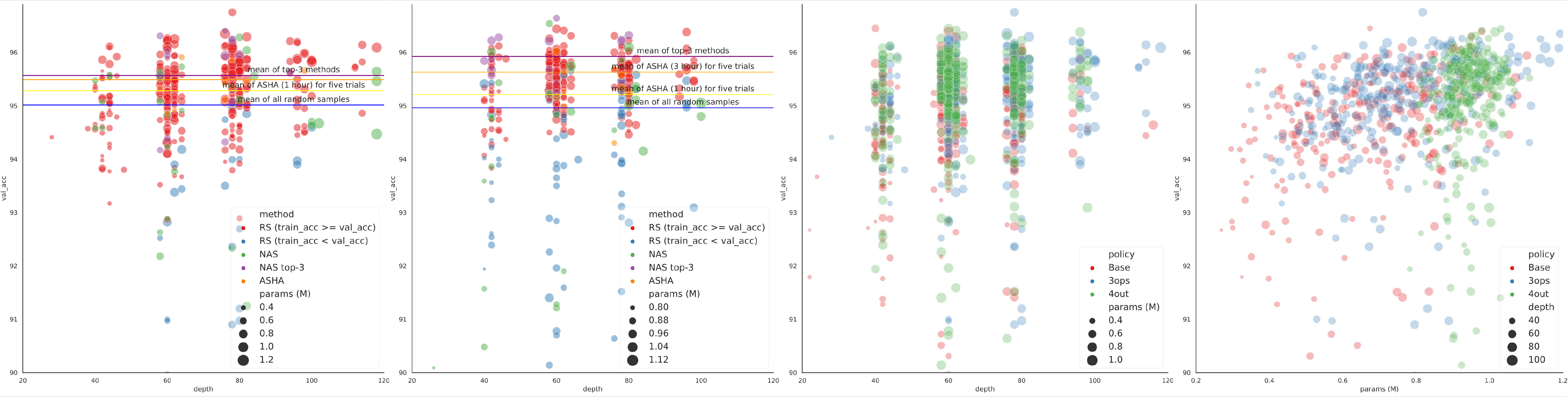}}
\caption{Random sampling and random search on C10\&3ops (i) and C10\&4out (ii). (iii) and (iiii) illustrate that \textit{val\_acc} is weakly correlated with depth and \textit{\#param} in all three valid spaces of LHD.}
\label{fig_rd_all}
\end{figure}

\begin{figure}[t]
\centerline{\includegraphics[width=0.4\textwidth]{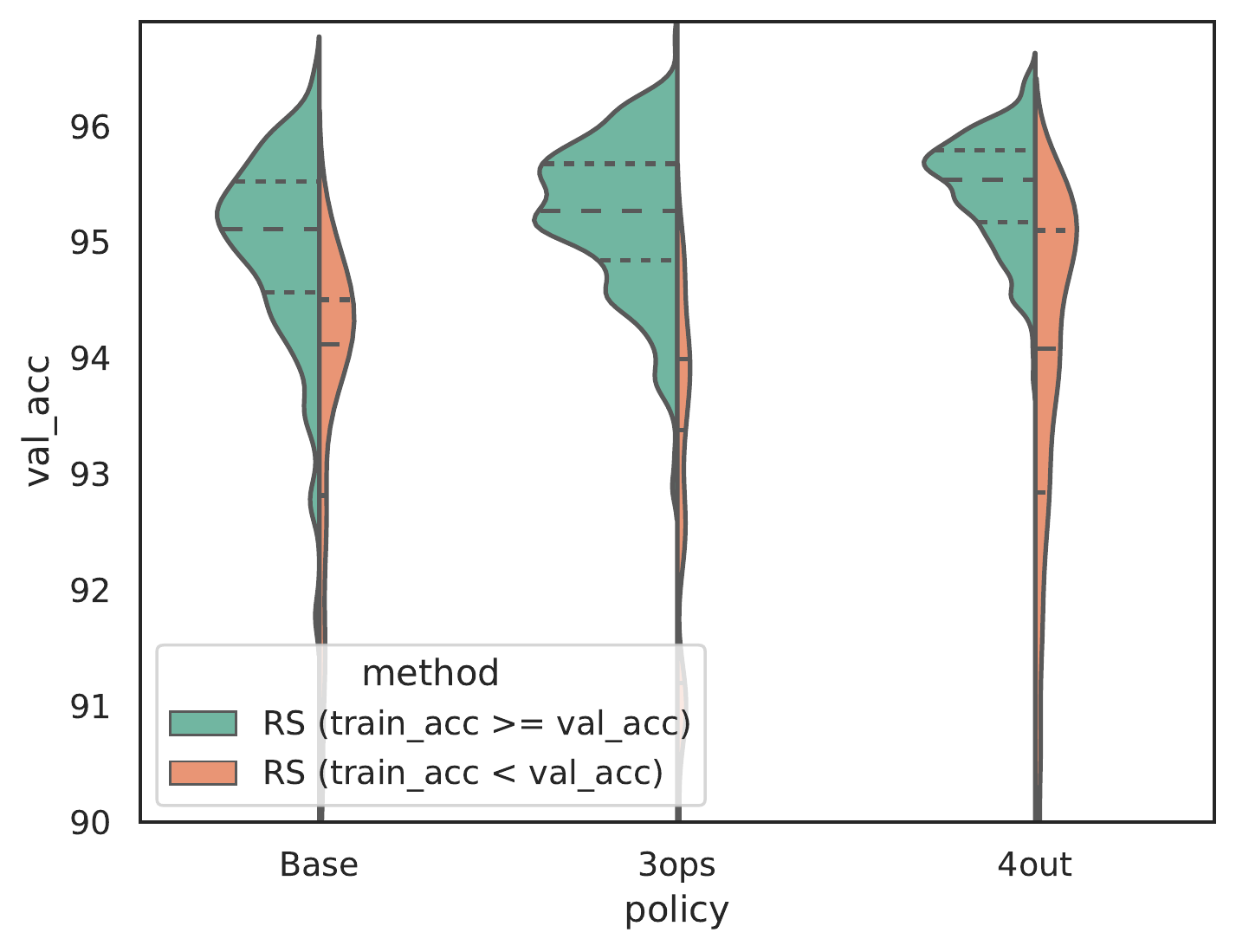}}
\caption{Base and 4out have a wider range of \textit{val\_acc} with a considerable proportion of hard-to-train samples. In comparison, 3ops can largely avoid failure cases and hard-to-train samples but the overall \textit{val\_acc} range is narrower.}
\label{fig_hard}
\end{figure}
\begin{figure}[t]
\centerline{\includegraphics[width=0.4\textwidth]{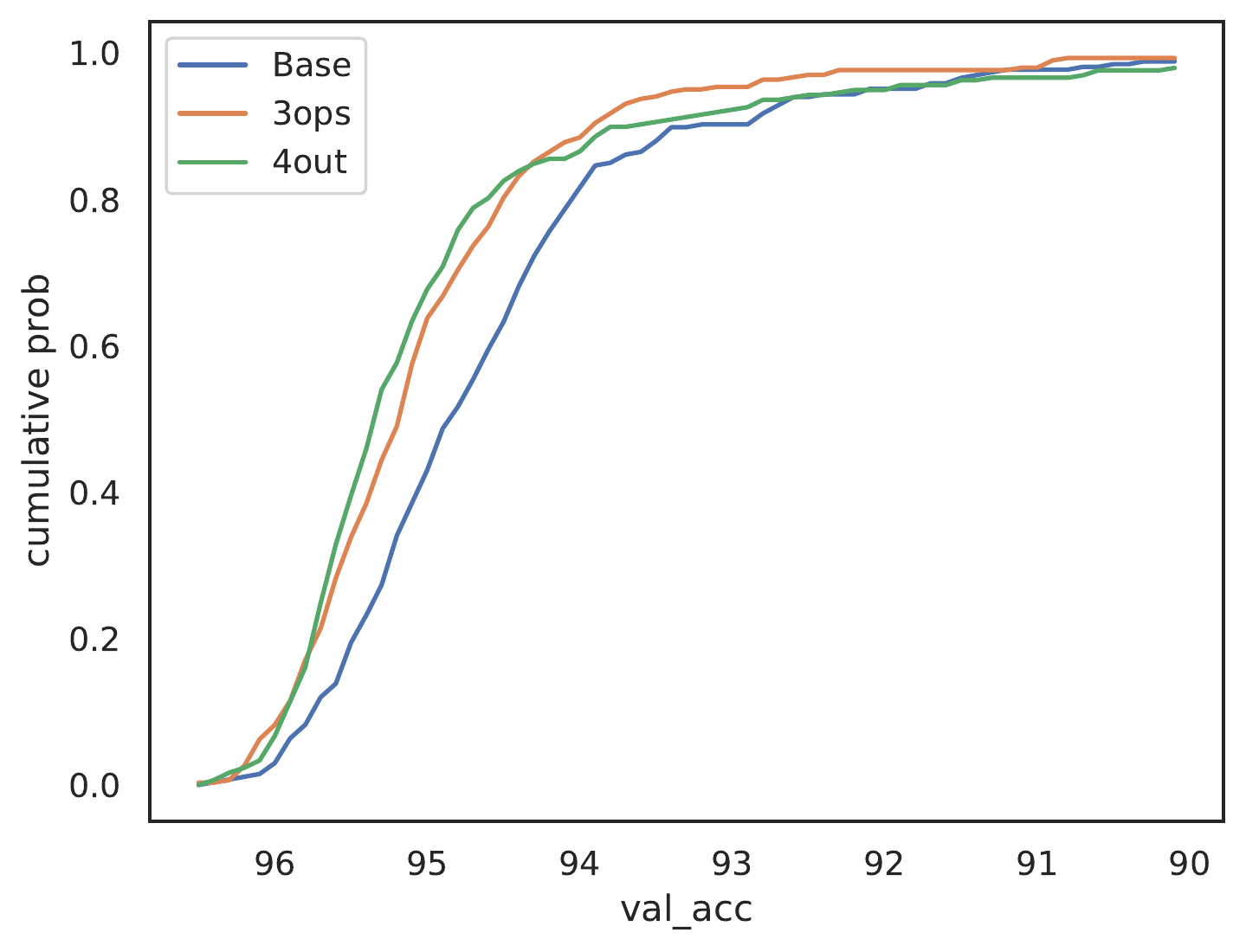}}
\caption{EDFs manifest the differences between search spaces through the curve gaps of the cumulative probability versus \textit{val\_acc} over random samples. 4out is close to 3out when the \textit{val\_acc} is $>$94\%, and close to Base when the \textit{val\_acc} is $<$93\%. We refer to \cite{radosavovic2019network} for more information about EDFs.}
\label{fig_edf}
\end{figure}

\section{Random sampling and Random search}
\citet{li2018massively} showed ASHA to be a state-of-the-art, theoretically principled, bandit-based partial training method that outperforms leading adaptive search strategies for hyperparameter optimization. \citet{li2020random} demonstrated when implemented properly, ASHA-based random search can deliver fairly competitive baselines against NAS methods after aligning the search cost. Our experiments are based on the codebase released by \citet{li2020random}\footnote{https://github.com/liamcli/randomNAS\_release} where we run ASHA with a starting resource per architecture of $r=1$ epoch and a maximum resource of 100 epochs w.r.t a promotion rate of $\eta=4$ which indicating the top-$\frac{1}{4}$ of architectures will be promoted in each round and trained for 4$\times$ more resources. We refer to \citet{li2020random} for more details of the random search. 

The results of random sampling (RS) and random search (ASHA) on C10\&3ops and C10\&4out are provide in Figure~\ref{fig_rd_all}(i) and (ii). We also illustrate all random samples on \textit{val\_acc} versus depth coordination in Figure~\ref{fig_rd_all}(iii) and \textit{val\_acc} versus \textit{\#param} coordination in Figure~\ref{fig_rd_all}(iiii) respectively. Figure~\ref{fig_hard} exhibits 
the proportion of sample accuracy distribution in different search spaces in which hard-to-train samples (\textit{train\_acc}$<$\textit{val\_acc}) are particularly identified. \citet{radosavovic2019network} proposed to characterize the distributions of architecture spaces through empirical distribution functions (EDFs) in a cumulative probability versus \textit{val\_acc} coordination as shown in Figure~\ref{fig_edf}.

\end{document}